\documentclass[10pt,twocolumn,letterpaper]{article}

\usepackage[pagenumbers]{cvpr} 

\usepackage[utf8]{inputenc}
\usepackage[T1]{fontenc}

\definecolor{cvprblue}{rgb}{0.21,0.49,0.74}
\definecolor{bestgreen}{RGB}{227, 252, 227}
\usepackage[pagebackref,breaklinks,colorlinks,allcolors=cvprblue]{hyperref}
\usepackage{booktabs}
\usepackage{multirow}
\usepackage[table]{xcolor}
\usepackage{colortbl}
\usepackage{dsfont}
\usepackage{placeins}
\usepackage{algorithm}
\usepackage{algpseudocode}
\definecolor{lightblue}{RGB}{225,232,240}
\newcommand{\our}[1]{\cellcolor{lightblue}#1}

\def\confName{CVPR}
\def\confYear{2027}

\title{GeArt: Global Explanation-Based Articulated Object Reconstruction}

\author{
Haiyan Sun$^{1*}$ \quad Dan Wang$^{2*}$ \quad Xinrui Cui$^{1}$ \\
$^{1}$University of North Texas \quad $^{2}$University of Arizona \\
}

\begin{document}
\maketitle

{
\renewcommand{\thefootnote}{\fnsymbol{footnote}}
\footnotetext[1]{Equal contribution.}
}

\begin{abstract}
Interactive digital twins of articulated objects are essential for robots to understand and manipulate everyday environments, yet reconstructing them requires recovering geometry, part segmentation, and joint parameters. Existing methods rely on non-convex optimization with unreliable motion initialization, where errors in geometry, segmentation, and motion can compensate for each other, leading to local minima, especially for complex multi-part objects. We present GeArt, a global explanation-based framework that infers the articulation structure through competition among motion hypotheses, providing a reliable initialization that explains both observed states for subsequent rendering-based refinement. To obtain reliable motion evidence, surface-aware dynamic detection leverages observed surfaces and cross-state free-space evidence to identify moving Gaussians. GeArt then generates revolute and prismatic joint hypotheses from local fragments of the detected dynamic Gaussians and selects a consistent motion set by minimizing a global energy over geometric agreement, appearance agreement, and kinematic plausibility. Using the inferred articulation, a mid-state canonical model is constructed and further refined through rendering-based optimization under multi-view RGB-D supervision. Experiments on synthetic and real-world datasets show that GeArt markedly outperforms existing methods in joint estimation and part-level geometry on complex multi-part and challenging real-world objects.
\end{abstract}

\vspace{-10pt}
\section{Introduction}
Articulated objects, such as cabinets, ovens, and laptops, consist of multiple rigid parts connected by joints that define their kinematic structure. With the rapid advancement of robotics and embodied AI, articulated object reconstruction has become increasingly important for enabling robots to understand and manipulate these everyday objects~\cite{wu2021vat,ma2023sim2real,geng2023gapartnet,mo2021where2act,xu2022universal}. Reconstructing articulated objects requires not only recovering their 3D geometry but also segmenting individual parts and estimating their joint types and kinematic parameters.

\begin{figure}[t]
\centering
\includegraphics[width=1\linewidth]{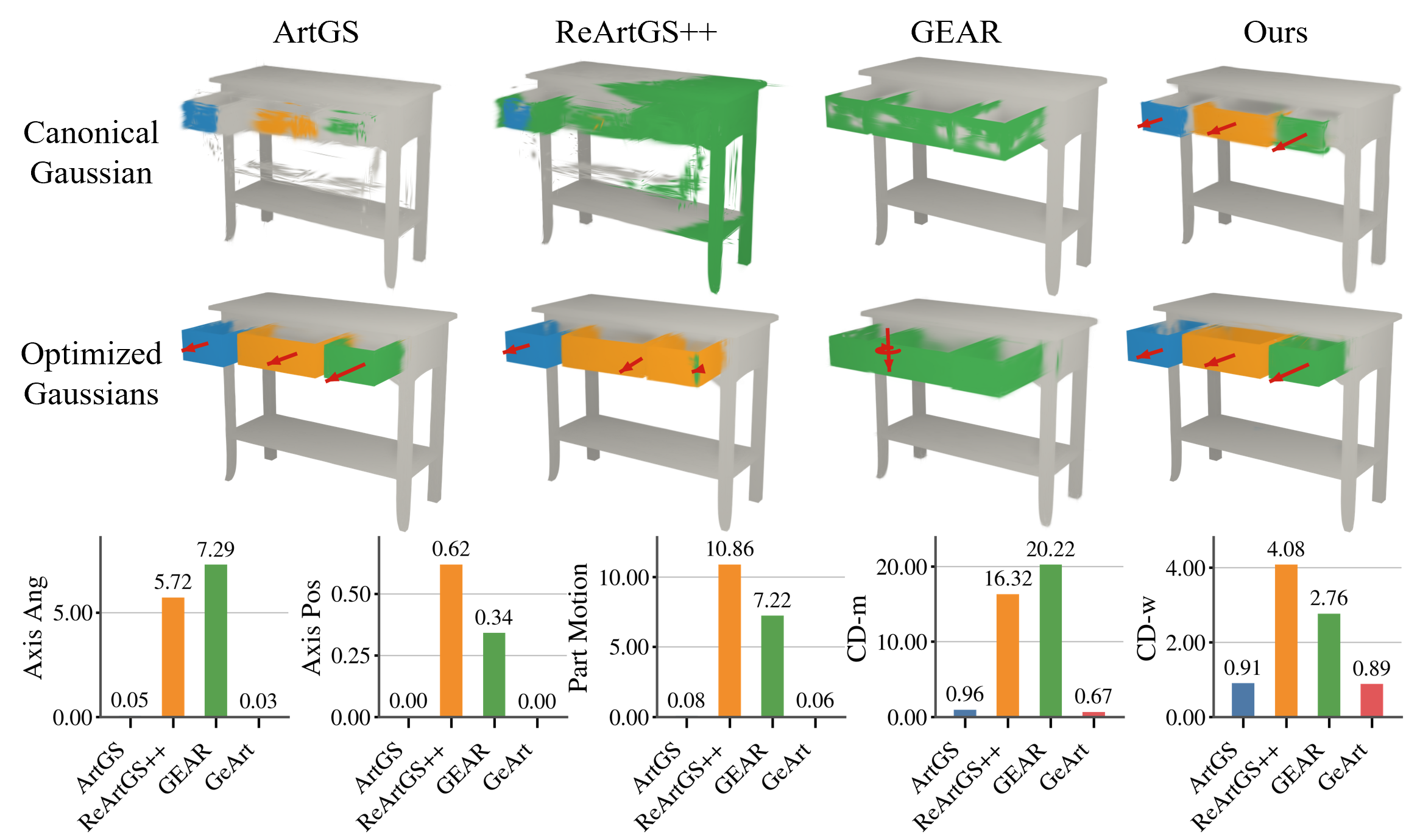}
\caption{Impact of canonical Gaussian initialization on articulated reconstruction. Existing methods often produce unreliable canonical Gaussians for objects with similar parts before refinement, leading to large errors in the optimized Gaussians. In contrast, GeArt uses global articulation inference to obtain a high-quality canonical Gaussian initialization, which enables more accurate geometry reconstruction and joint estimation after optimization.}
\label{fig:teaser}
\end{figure}

Recent methods leverage 3D Gaussian Splatting (3DGS)~\cite{kerbl20233d} to reconstruct the geometry of articulated objects, achieving promising visual fidelity. However, accurately recovering their articulation parameters remains challenging. For discrete structural variables, such as part assignments and joint types, some methods rely on pretrained segmentation models or assume predefined joint types, introducing additional dependencies and limiting their generality~\cite{shen2026gaussianart}. For continuous kinematic parameters, most existing methods optimize them jointly or progressively with geometry and part assignments through gradient-based optimization, combining rendering supervision with additional constraints and regularization~\cite{liu2025building,yu2026part,wu2026reartgs++}. However, such optimization is sensitive to initialization: when the true joint motion differs substantially from its initialization, or geometrically similar parts introduce ambiguous correspondences, it converges to incorrect motion estimates~\cite{lin2025splart,guo2025articulatedgs}. As shown in Fig.~\ref{fig:teaser}, an unreliable canonical initialization on an object with similar parts leads to large errors after optimization. The underlying cause is that geometry and motion are coupled under the rendering objective, so errors in one can be absorbed by the other: inaccurate motion can be compensated for by distorting the reconstructed geometry, and vice versa, reducing the rendering loss without recovering the correct articulation~\cite{li2026gear}. This coupling traps gradient-based optimization in poor local minima.

To address these challenges, we propose GeArt, a global explanation-based articulated object reconstruction framework that infers a coarse articulation structure by minimizing a global energy over competing motion hypotheses, providing a reliable initialization for subsequent gradient-based rendering optimization. As illustrated in Fig.~\ref{fig:pipeline}, we first introduce surface-aware dynamic Gaussian detection to identify moving surface Gaussians while filtering out interior and floating Gaussians that may be incorrectly classified as dynamic. We then cast articulation discovery as geometric multi-model fitting~\cite{isack2012energy}. Each model is a typed joint. The energy scores a joint set on both observed states with geometric term, appearance term, and kinematic plausibility. By minimizing this energy, GeArt jointly infers the coarse articulation structure, including part assignments, joint types, and initial joint parameters. This global explanation-based formulation resolves competing motion hypotheses under a shared objective, reducing ambiguities among geometrically similar parts and mitigating the coupled error compensation of gradient-based optimization. We build the canonical model at the mid state of the inferred joints, which provides a reliable initialization for subsequent refinement, where Gaussian geometry, part assignments, and joint parameters are further optimized using multi-view RGB-D rendering supervision.

Our contributions are summarized as follows:
\begin{itemize}
    \item We introduce \textbf{GeArt}, which infers the articulation structure through \textbf{global articulation inference} before rendering-based refinement. GeArt generates joint hypotheses from local fragments of the observed motion and selects the set that minimizes a global explanation energy over both states. The selected set gives the part assignments, joint types, and initial joint parameters for the subsequent refinement.

    \item We propose \textbf{surface-aware dynamic detection}, which identifies moving Gaussians from observed surfaces and cross-state free-space evidence, avoiding false positives from interior and floating Gaussians in conventional nearest-neighbor-based motion detection and providing clean motion evidence for the inference.

    \item We construct mid-state canonical gaussian that enables accurate articulation recovery and part segmentation through rendering-based optimization, ultimately improving performance on complex multi-part objects.
\end{itemize}

\section{Related Work}

\subsection{Articulated Object Reconstruction}
Articulated object reconstruction aims to recover rigid part geometry and the joints governing relative part motion. Early approaches estimated kinematic structures by fitting motion models to observed part poses or tracked 3D features, using relative motion to identify rigid components and their connections~\cite{sturm2011probabilistic,katz2013interactive}. These approaches established geometric motion consistency as a basis for discovering articulation from object interactions. Subsequent prediction-based methods learned shape-motion associations to infer part mobility from static geometry~\cite{hu2017learning,wang2019shape2motion,yan2020rpm}, while related approaches estimated part poses and joint parameters from image or depth observations~\cite{li2020category,jain2021screwnet,jiang2022opd}. Methods using paired shapes or interaction observations further exploited cross-state correspondences and deformation cues to infer part structure and motion~\cite{yi2018deep,jiang2022ditto}. Learned implicit representations extended this direction by encoding shape and articulation through signed distance functions, occupancy fields, or radiance fields, supporting articulated reconstruction through prediction or latent-variable fitting~\cite{mu2021sdf,kawana2022unsupervised,tseng2022cla,heppert2023carto}. These methods exploit reusable shape and articulation priors, but their coverage depends on the training data, and many require 3D assets or part and joint annotations that are costly to obtain.

\subsection{Dynamic Gaussian Modeling}
3D Gaussian Splatting provides an explicit scene representation with efficient differentiable rendering~\cite{kerbl20233d}. Early dynamic extensions model scene evolution through persistent Gaussian trajectories, canonical-space deformation fields, or time-dependent Gaussian attributes, enabling reconstruction and novel-view synthesis from multi-view or monocular videos~\cite{luiten2024dynamic,yang2024deformable,wu20244d,duisterhof2023md}. More structured representations use sparse control points or shared trajectory bases to coordinate Gaussian motion and support editing~\cite{huang2024sc,kratimenos2024dynmf}. These methods capture flexible scene dynamics, but their motion variables are generally not parameterized as the revolute or prismatic joints governing rigid articulated objects.

Recent articulated Gaussian methods associate primitives with rigid parts and shared transformations or screw motions, enabling geometry, part segmentation, and motion to be estimated from observations of different articulation states~\cite{liu2025building,guo2025articulatedgs,kim2025screwsplat,shen2026gaussianart}. Their common principle is to explain cross-state appearance changes through coordinated transformations of part-associated Gaussians. Additional geometric and kinematic constraints improve surface reconstruction and motion consistency, including SDF-based surface guidance and constraints on part contact, motion, and separation~\cite{wu2026reartgs,yu2026part}. Existing methods that jointly optimize Gaussian geometry, part assignments, and motion suffer from tightly coupled estimation: errors in geometry or segmentation can be compensated for by inaccurate motion estimates, and vice versa. Consequently, optimization can converge to an incorrect articulation model despite low rendering error. To overcome these challenges, we select a set of motion hypotheses that collectively explains both observed states, recovering coarse part assignments, joint types, and initial joint parameters before rendering-based optimization. This inferred structure provides a reliable initialization for jointly refining Gaussian geometry, part segmentation, and joint parameters under RGB-D supervision.
\section{Method}
\label{sec:method}

\begin{figure*}[!t]
  \centering
  \includegraphics[width=1\textwidth]{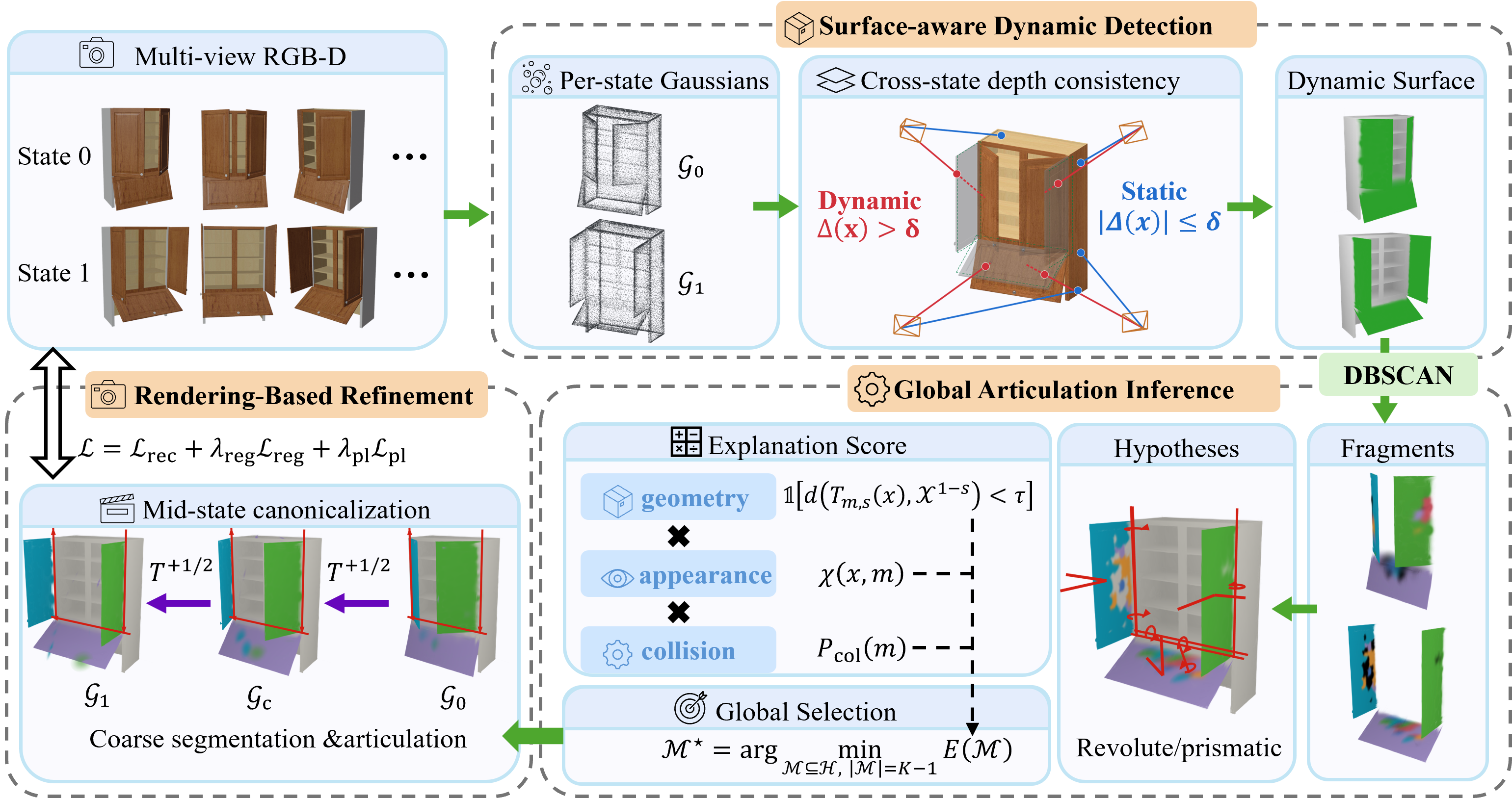}
  \vspace{-10pt}
  \caption{\textbf{Overview of GeArt.} Given two-state multi-view RGB-D observations, per-state Gaussians are reconstructed, and surface-aware dynamic detection identifies moving Gaussians. Local fragments of the moving Gaussians are used to generate diverse revolute and prismatic joint hypotheses. A global explanation score combining geometry, appearance, and kinematic plausibility selects $K-1$ joint hypotheses, yielding coarse part segmentation, joint types, and initial joint parameters. These estimates are further refined through rendering-based optimization.}
  \label{fig:pipeline}
  \vspace{-10pt}
\end{figure*}

\subsection{Framework Overview}
\label{sec:method_overview}

\paragraph{Problem setting.}
Given calibrated multi-view RGB-D images of an articulated object in two states $s\in\{0,1\}$, along with the number of parts $K$, our objective is to recover its geometry, part segmentation, and estimate joint parameters. The object consists of a static base and $K-1$ rigid movable parts, each connected to the base by a revolute or prismatic joint. We formulate articulated reconstruction as a mixed discrete-continuous problem, where the part assignment of each Gaussian and the joint type are discrete variables, while the Gaussian geometry and joint parameters $\boldsymbol\Theta_k$ are continuous variables. We explicitly model the motion of each joint from state 0 to state 1 as
\begin{equation}
    T_k(\mathbf x)=
    \begin{cases}
        \mathbf R(\mathbf a_k,\theta_k)(\mathbf x-\mathbf p_k)+\mathbf p_k, & \text{revolute},\\
        \mathbf x+s_k\mathbf u_k, & \text{prismatic},
    \end{cases}
    \label{eq:joint}
\end{equation}
where $\mathbf a_k$, $\mathbf p_k$, and $\theta_k$ denote the axis direction, pivot point, and rotation angle of a revolute joint, respectively, while $\mathbf u_k$ and $s_k$ denote the translation direction and displacement of a prismatic joint, respectively. 

\vspace{-10pt}
\paragraph{Overview.}
As shown in Fig.~\ref{fig:pipeline}, GeArt reconstructs each state independently and applies surface-aware dynamic detection (Sec.~\ref{sec:detection}) to identify dynamic Gaussians. Global articulation inference (Sec.~\ref{sec:Global}) fits revolute and prismatic hypotheses to fragments of these Gaussians and selects the $K-1$ joints that jointly explain them in both states, yielding initial part labels $\boldsymbol\ell^{(0)}$, joint types $\boldsymbol\kappa$, and joint parameters $\boldsymbol\Theta^{(0)}$. Rendering-based refinement (Sec.~\ref{sec:refine}) then builds mid-state canonical Gaussians from this structure and jointly refines geometry, part segmentation, and joint parameters under RGB-D supervision.

\subsection{Surface-aware Dynamic Detection}
\label{sec:detection}
Prior approaches distinguish dynamic from static Gaussians using simple nearest-neighbor distance comparisons between independently reconstructed states~\cite{liu2025building,wu2026reartgs++,yu2026part}, but they often misclassify interior or floating Gaussians as dynamic due to cross-state reconstruction inconsistencies, leading to inaccurate initialization of dynamic Gaussians. To address this limitation, we propose a surface-aware dynamic Gaussian detection method that identifies dynamic Gaussians based on observed surfaces and cross-state free-space evidence. For each state $s$, we reconstruct
\begin{equation}
    \mathcal G=\{g_i\}_{i=1}^{N},\qquad
    g_i=(\boldsymbol\mu_i,\boldsymbol\Sigma_i,\alpha_i,\mathbf c_i),
    \label{eq:gaussian}
\end{equation}
where $\boldsymbol\mu_i$, $\boldsymbol\Sigma_i$, $\alpha_i$, and $\mathbf c_i$ denote the center, covariance, opacity, and appearance of each Gaussian, respectively. Following differentiable Gaussian rasterization, we optimize each state using multi-view RGB-D supervision:
\begin{equation}
\begin{aligned}
\mathcal L_{\mathrm{rec}}
=&\ \lambda_1\|I-\widehat I\|_1
+\lambda_{\mathrm{SSIM}}\bigl(1-\operatorname{SSIM}(I,\widehat I)\bigr)\\
&+\lambda_D\|D-\widehat D\|_1,
\end{aligned}
\label{eq:RGB-D_loss}
\end{equation}
where $\{I,D\}$ and $\{\widehat I,\widehat D\}$ denote the observed and rendered RGB-D images, respectively.

Intuitively, articulated motion changes the space occupied by an object, which can be observed through depth maps. Although the two states are reconstructed independently, they share a common world coordinate system, allowing Gaussians from either state to be directly projected into the other state's camera views. Static surface Gaussians lie on observed surfaces in both states, whereas dynamic ones lie on observed surfaces in their own state and in free space observed by the other state.
When a Gaussian's location is occluded by a closer surface in the other state, for instance by a drawer opened in front of it, the view cannot determine whether that location is still occupied and abstains.

\begin{figure}[t]
  \centering
  \includegraphics[width=\linewidth]{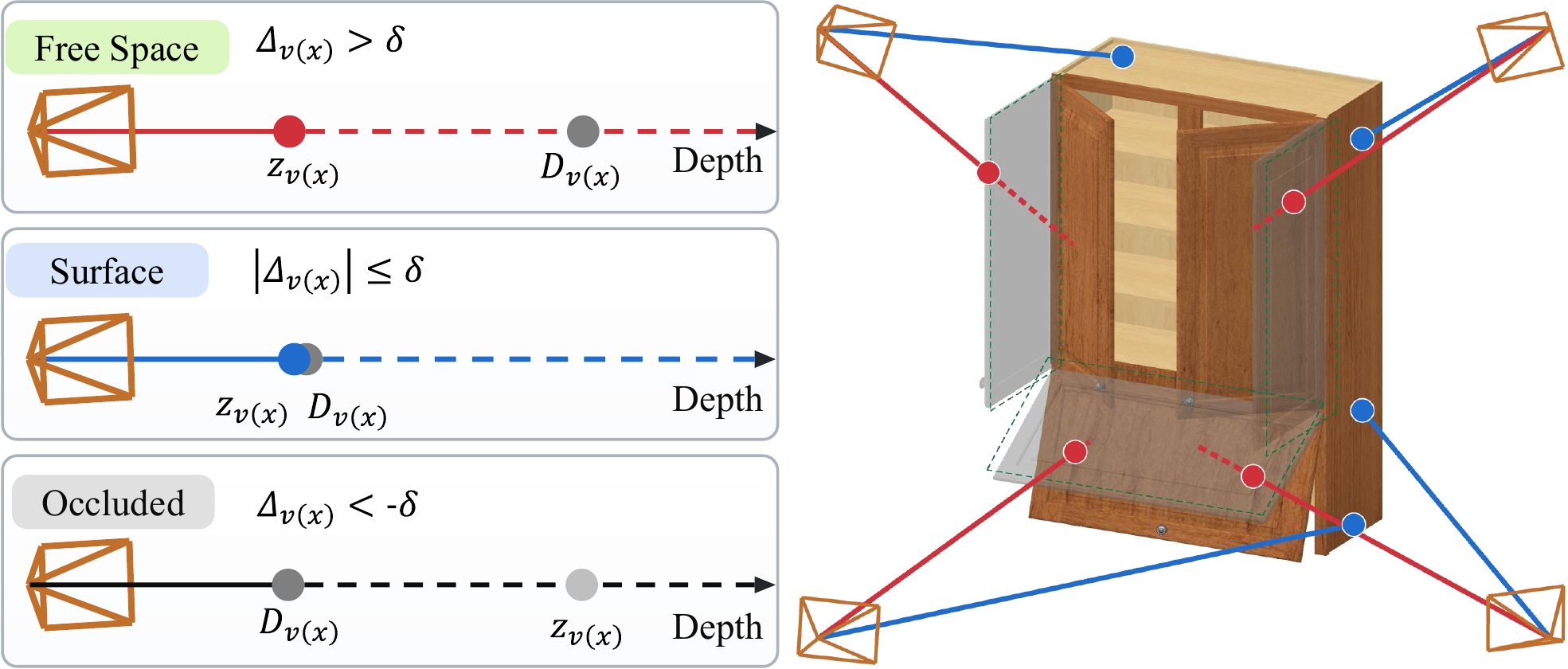}
  \caption{A Gaussian on an observed surface of its own state is projected into the views of the other state and compared with the observed depth. Free space marks it as dynamic, a matching surface marks it as static, and occluded views are ignored.}
  \label{fig:dynamic}
  \vspace{-10pt}
\end{figure}

As shown in Fig.~\ref{fig:dynamic}, for a Gaussian center $\mathbf x$ projected into a view $v$ of either state, we define the depth difference
\begin{equation}
    \Delta_v(\mathbf x)=D_v(\mathbf x)-z_v(\mathbf x),
\end{equation}
where $D_v(\mathbf x)$ is the observed depth at the projected pixel and $z_v(\mathbf x)$ is the Gaussian's depth in that camera. Pixels without valid depth, such as the background, are treated as infinitely far. A difference above $\delta$ indicates free space at $\mathbf x$, a difference within $\pm\delta$ indicates a surface at $\mathbf x$, and a difference below $-\delta$ indicates an occluder in front of $\mathbf x$.

We identify dynamic Gaussians by combining surface consistency, cross-state geometric separation, and multi-view free-space evidence. Let $\mathcal X^s$ denote the centers of Gaussians in state $s$, and $\mathcal V^s$ denote the camera views of state $s$. The detected dynamic set is
\begin{equation}
\begin{aligned}
\mathcal{D}^s=\bigl\{\mathbf x\in\mathcal X^s:\;
&\min_{v\in\mathcal V^s}|\Delta_v(\mathbf x)|\le\delta,\\
&d(\mathbf x,\mathcal X^{1-s})>\epsilon,\\
&n_{\mathrm{free}}(\mathbf x)>\max(n_{\mathrm{surf}}(\mathbf x),1)
\bigr\},
\end{aligned}
\label{eq:detect}
\end{equation}
where $d(\cdot,\cdot)$ is the nearest-neighbor distance, and $n_{\mathrm{free}}$ and $n_{\mathrm{surf}}$ count the views in $\mathcal V^{1-s}$ with $\Delta_v>\delta$ and $|\Delta_v|\le\delta$, respectively; occluded views enter neither count. Both $\delta$ and $\epsilon$ are scaled to the object size. We apply this detection bidirectionally to obtain dynamic Gaussians in both states. 

\subsection{Global Articulation Inference}
\label{sec:Global}
The surface-aware detection provides high-confidence dynamic Gaussians in both states, but does not determine their part assignments or joint motions. Existing methods often initialize coarse part assignments through clustering and subsequently jointly optimize part assignments and joint parameters. However, such initialization is unreliable when movable parts are spatially adjacent or geometrically similar, such as stacked drawers, where clustering may incorrectly group distinct parts. Because part assignments and joint parameters are highly coupled, errors in one can be compensated by the other, potentially leading to poor local minima. To obtain a more robust initialization, we propose an energy-based global articulation inference method that generates diverse typed joint hypotheses from local Gaussian fragments and globally selects the set that best explains the dynamic Gaussians across both states. 

\vspace{-5pt}
\paragraph{Joint Hypothesis Generation.}
We generate a diverse pool of candidate joint motions from local fragments of the dynamic Gaussians in both states. DBSCAN~\cite{ester1996density} first partitions the dynamic Gaussians of each state into small fragments, and every pair of similarly sized fragments from the two states serves as a candidate correspondence. For each pair, candidate axis directions come from the fragment and the static base, and the rotation angle or the displacement is swept along each direction. The pivot of a revolute candidate follows in closed form from the condition that the rotation maps one fragment centroid onto the other. The best candidates are then refined with iterative closest point~\cite{besl1992method}, restricted to the angle and pivot for a revolute joint and to the translation for a prismatic joint. Since fragments of the same part produce nearly identical hypotheses, we group hypotheses that move the Gaussians they explain to nearly the same positions. All hypotheses form the candidate pool $\mathcal{H}$ for the subsequent global joint-set selection.

\vspace{-5pt}
\paragraph{Explanation Score.}
We first measure how well a single hypothesis $m\in\mathcal{H}$ explains a single dynamic Gaussian. Let $T_{m,s}$ denote the motion of hypothesis $m$ from state $s$ to the other state, and let $r_m(\mathbf{x}) = d\bigl(T_{m,s}(\mathbf{x}),\mathcal{X}^{1-s}\bigr)$
be the residual of $\mathbf{x}\in\mathcal{D}^s$ under $m$. A joint $m$ explains $\mathbf{x}$ when it moves $\mathbf{x}$ onto the surface of the other state,
\begin{equation}
c(\mathbf{x},m)=\tfrac{1}{2}\,\mathbf{1}\bigl[r_m(\mathbf{x})<\tau\bigr]
\bigl(1+\chi(\mathbf{x},m)\bigr),
\label{eq:observation_term}
\end{equation}
where $\mathcal X^{1-s}$ denotes the Gaussian centers reconstructed of the other state, $\chi(\mathbf{x},m)\in\{0,1\}$ indicates appearance agreement with the nearest neighbor, and $\tau$ is $1\%$ of the bounding-box diagonal. A physical joint moves its part without passing through static geometry. We test for such collisions by proximity. Let $\mathcal X_{\mathrm{stat}}$ be the union of the static Gaussian centers of the two states. A point is in contact with static geometry when its nearest center in $\mathcal X_{\mathrm{stat}}$ is closer than $\tau$. Let $S_m$ be the sampled centers of dynamic Gaussians that support hypothesis $m$. The collision ratio averages the fraction of new contacts over three intermediate poses,
\begin{equation}
\begin{aligned}
P_{\mathrm{col}}(m)
={}&\frac{1}{|\mathcal A||S_m|}\sum_{\alpha\in\mathcal A}\sum_{\mathbf x\in S_m}\\
&\mathbf{1}\bigl[d(\mathbf x,\mathcal X_{\mathrm{stat}})\ge\tau\bigr]\\
&{}\times\mathbf{1}\bigl[d(T_m^{\alpha}(\mathbf x),\mathcal X_{\mathrm{stat}})<\tau\bigr],
\end{aligned}
\label{eq:collision_ratio}
\end{equation}
where $\mathcal{A}=\{1/4,1/2,3/4\}$ denotes the sampled intermediate articulation states, and $T_m^\alpha$ applies the fraction $\alpha$ of the joint angle or displacement. The indicator $\mathbf{1}\bigl[d(\mathbf x,\mathcal X_{\mathrm{stat}})\ge\tau\bigr]$ excludes points that are already in contact at rest, such as the edge of a door next to its frame. The kinematic plausibility lowers the score of a hypothesis in proportion to this ratio,
\begin{equation}
\pi(m)=1-\tfrac{1}{2}P_{\mathrm{col}}(m).
\label{eq:kinematic_prior}
\end{equation}
The explanation score combines geometric and appearance agreement with the kinematic plausibility:
\begin{equation}
w(\mathbf{x},m)=\pi(m)\,c(\mathbf{x},m).
\label{eq:explanation_score}
\end{equation}

\vspace{-10pt}
\paragraph{Global Joint-set Selection.}
A candidate structure is a set $\mathcal M\subseteq\mathcal H$ of $K-1$ joints. Geometry decides which joint moves each Gaussian. We assign each dynamic Gaussian $\mathbf x$ to the joint with the smallest residual $r_m(\mathbf{x})$. With geometrically similar parts, several joints can move $\mathbf x$ to within $\tau$ of the other state. In this case, we assign $\mathbf{x}$ to the joint that moves it by the smaller displacement. We denote the resulting label by $\ell_{\mathcal M}(\mathbf x)$. Each Gaussian then contributes to the energy through its explanation score under the assigned joint. We score $\mathcal M$ by the motion evidence that it leaves unexplained in both states,
\begin{equation}
E(\mathcal M)=\sum_{s=0}^{1}\sum_{\mathbf x\in\mathcal{D}^s}\Bigl[1-w\bigl(\mathbf x,\ell_{\mathcal M}(\mathbf x)\bigr)\Bigr].
\label{eq:global_energy}
\end{equation}
We select the set with the lowest energy,
\begin{equation}
\mathcal M^{\star}=\arg\min_{\mathcal M\subseteq\mathcal H,\;|\mathcal M|=K-1}E(\mathcal M).
\label{eq:joint_selection}
\end{equation}
Each Gaussian contributes to $E$ once, through its assigned joint. Two hypotheses of the same part thus share its Gaussians, while a hypothesis of another part explains Gaussians that are otherwise unexplained. The energy therefore favors sets whose joints explain different parts. We solve Eq.~\eqref{eq:joint_selection} with a greedy construction followed by the interchange heuristic~\cite{teitz1968heuristic}. Starting from one representative per group, we add the joint that lowers $E$ most until the set has $K-1$ joints. We then swap one selected joint for other hypotheses of $\mathcal{H}$ whenever the swap lowers $E$. After the swaps, we fit new hypotheses to the Gaussians assigned to each selected joint, add them to $\mathcal{H}$, and repeat the swaps until the selected joints no longer change.

The selected set $\mathcal{M}^\star$ determines the coarse articulation structure. Its $K-1$ joints give the joint types $\boldsymbol{\kappa}$ and the initial joint parameters $\boldsymbol{\Theta}^{(0)}$. The initial part labels $\boldsymbol{\ell}^{(0)}$ assign each dynamic Gaussian $\mathbf{x}$ to the part of joint $\ell_{\mathcal{M}^\star}(\mathbf{x})$. We then assign remaining Gaussians to parts whose transformed dynamic Gaussians from the other state lie nearby. All other Gaussians form the static base.

\begin{table*}[t]
\centering
\begingroup
\caption{\textbf{Quantitative results on \textsc{\textbf{Paris}}}. Lower ($\downarrow$) is better across all metrics. \textbf{Best}, \underline{second}. ``-'' indicates objects containing only prismatic joints.}
\vspace{-5pt}
\label{tab:paris_result}
\resizebox{\linewidth}{!}{
\begin{tabular}{cc|cccccccccc|cc|c}
\hline
\multirow{2}{*}{Metric} & \multirow{2}{*}{Method} & \multicolumn{10}{c}{Synthetic Objects} & \multicolumn{2}{|c}{Real Objects} & \multicolumn{1}{c}{} \\
& & FoldChair & Fridge & Laptop & Oven & Scissor & Stapler & USB & Washer & Blade & Storage & Fridge & Storage & Average \\
\hline
\multirow{4}{*}{\shortstack{Axis\\Ang}} & ArtGS & \textbf{0.00} & 0.03 & 0.05 & \underline{0.02} & 0.15 & \textbf{0.00} & 0.07 & \underline{0.04} & 0.11 & 0.05 & 2.05 & \underline{3.65} & \underline{0.52} \\
& ReArtGS++ & \underline{0.01} & \underline{0.02} & \underline{0.01} & \textbf{0.00} & \underline{0.06} & \underline{0.01} & \textbf{0.05} & \textbf{0.03} & 1.27 & 0.04 & \textbf{1.87} & 16.56 & 1.66 \\
& GEAR & \textbf{0.00} & \textbf{0.00} & 0.03 & \textbf{0.00} & \textbf{0.00} & \textbf{0.00} & 0.31 & \textbf{0.03} & \underline{0.04} & \underline{0.03} & 2.90 & 70.59 & 6.16 \\
& \our{\textbf{Ours}} & \our{0.03} & \our{0.03} & \our{\textbf{0.00}} & \our{\textbf{0.00}} & \our{\underline{0.06}} & \our{0.02} & \our{\underline{0.06}} & \our{\textbf{0.03}} & \our{\textbf{0.03}} & \our{\textbf{0.00}} & \our{\underline{2.03}} & \our{\textbf{1.97}} & \our{\textbf{0.36}} \\
\hline
\multirow{4}{*}{\shortstack{Axis\\Pos}} & ArtGS & \underline{0.01} & \textbf{0.00} & \textbf{0.01} & \textbf{0.00} & \textbf{0.00} & \textbf{0.01} & \textbf{0.00} & \textbf{0.00} & - & - & 0.46 & - & \underline{0.05} \\
& ReArtGS++ & \textbf{0.00} & \textbf{0.00} & \textbf{0.01} & \textbf{0.00} & \textbf{0.00} & \textbf{0.01} & \textbf{0.00} & \underline{0.01} & - & - & \underline{0.42} & - & \textbf{0.05} \\
& GEAR & \textbf{0.00} & \textbf{0.00} & \underline{0.02} & \textbf{0.00} & \textbf{0.00} & \underline{0.02} & \underline{0.67} & \underline{0.01} & - & - & \textbf{0.32} & - & 0.12 \\
& \our{\textbf{Ours}} & \our{\textbf{0.00}} & \our{\underline{0.01}} & \our{\textbf{0.01}} & \our{\underline{0.01}} & \our{\textbf{0.00}} & \our{\textbf{0.01}} & \our{\textbf{0.00}} & \our{\underline{0.01}} & \our{-} & \our{-} & \our{0.57} & \our{-} & \our{0.07} \\
\hline
\multirow{4}{*}{\shortstack{Part\\Motion}} & ArtGS & \underline{0.01} & 0.05 & 0.03 & \underline{0.02} & 0.10 & \textbf{0.00} & \underline{0.06} & \underline{0.03} & \textbf{0.00} & \textbf{0.00} & \underline{1.93} & \underline{0.04} & \underline{0.19} \\
& ReArtGS++ & \textbf{0.00} & \textbf{0.02} & \textbf{0.01} & \textbf{0.01} & \underline{0.03} & \underline{0.02} & \textbf{0.04} & \textbf{0.02} & \textbf{0.00} & \textbf{0.00} & 2.30 & 0.25 & 0.23 \\
& GEAR & 0.02 & \textbf{0.02} & 0.04 & \underline{0.02} & \textbf{0.00} & 0.05 & 1.31 & \underline{0.03} & \textbf{0.00} & \textbf{0.00} & 2.85 & 0.31 & 0.39 \\
& \our{\textbf{Ours}} & \our{0.04} & \our{\underline{0.04}} & \our{\underline{0.02}} & \our{\textbf{0.01}} & \our{0.04} & \our{\underline{0.02}} & \our{\textbf{0.04}} & \our{\underline{0.03}} & \our{\textbf{0.00}} & \our{\textbf{0.00}} & \our{\textbf{1.77}} & \our{\textbf{0.03}} & \our{\textbf{0.17}} \\
\hline
\multirow{4}{*}{CD-s} & ArtGS & 6.23 & \underline{0.49} & 0.62 & \underline{3.90} & 0.47 & \underline{3.82} & 2.38 & 6.46 & \textbf{0.50} & 7.38 & \textbf{1.75} & 4.35 & 3.20 \\
& ReArtGS++ & 0.25 & 1.53 & 0.80 & 11.37 & 0.74 & 4.38 & 2.02 & 25.08 & 0.70 & 9.27 & 3.27 & 13.38 & 6.07 \\
& GEAR & \underline{0.14} & \textbf{0.47} & \textbf{0.42} & 4.00 & \underline{0.44} & \textbf{3.77} & \underline{1.93} & \textbf{5.08} & 1.32 & \underline{4.61} & \underline{1.87} & \underline{4.05} & \textbf{2.34} \\
& \our{\textbf{Ours}} & \our{\textbf{0.13}} & \our{0.61} & \our{\underline{0.60}} & \our{\textbf{3.77}} & \our{\textbf{0.43}} & \our{4.06} & \our{\textbf{1.77}} & \our{\underline{6.37}} & \our{\underline{0.62}} & \our{\textbf{4.18}} & \our{3.58} & \our{\textbf{2.23}} & \our{\underline{2.36}} \\
\hline
\multirow{4}{*}{CD-m} & ArtGS & \underline{0.31} & \textbf{0.21} & \underline{0.14} & \underline{0.38} & 0.48 & \underline{0.82} & 1.24 & \underline{0.22} & \underline{1.16} & \underline{0.93} & \textbf{0.93} & \underline{52.59} & \textbf{4.95} \\
& ReArtGS++ & 0.76 & 1.27 & 0.61 & 1.20 & 7.92 & 1.75 & \underline{1.21} & 0.96 & 28.11 & 4.72 & 2.25 & 278.63 & 27.45 \\
& GEAR & \textbf{0.30} & 0.32 & \textbf{0.11} & 0.72 & \textbf{0.39} & \textbf{0.55} & 2.24 & \textbf{0.08} & 168.75 & \textbf{0.72} & 20.32 & 204.02 & 33.21 \\
& \our{\textbf{Ours}} & \our{\underline{0.31}} & \our{\underline{0.29}} & \our{0.22} & \our{\textbf{0.36}} & \our{\underline{0.44}} & \our{1.13} & \our{\textbf{0.99}} & \our{0.37} & \our{\textbf{0.88}} & \our{5.75} & \our{\underline{1.03}} & \our{\textbf{48.60}} & \our{\underline{5.03}} \\
\hline
\multirow{4}{*}{CD-w} & ArtGS & 3.88 & \textbf{0.58} & \underline{0.49} & \underline{3.63} & 0.55 & \underline{2.87} & 1.56 & 6.19 & \underline{0.63} & 5.23 & \underline{1.19} & \underline{2.64} & \underline{2.45} \\
& ReArtGS++ & \underline{0.53} & \underline{1.76} & 0.84 & 10.38 & 4.14 & 3.37 & 1.56 & 23.48 & 0.72 & 7.38 & 3.00 & 5.07 & 5.19 \\
& GEAR & 11.59 & \textbf{0.58} & \textbf{0.45} & 4.00 & \textbf{0.48} & \textbf{2.75} & \underline{1.53} & \textbf{5.13} & 0.71 & \textbf{5.00} & \textbf{1.09} & 3.21 & 3.04 \\
& \our{\textbf{Ours}} & \our{\textbf{0.28}} & \our{\textbf{0.58}} & \our{0.62} & \our{\textbf{3.59}} & \our{\underline{0.54}} & \our{2.90} & \our{\textbf{1.39}} & \our{\underline{5.66}} & \our{\textbf{0.61}} & \our{\underline{5.12}} & \our{2.19} & \our{\textbf{2.33}} & \our{\textbf{2.15}} \\
\hline
\end{tabular}
}
\endgroup
\vspace{-5pt}
\end{table*}

\subsection{Rendering-Based Refinement}
\label{sec:refine}
Following~\cite{liu2025building}, we refine the geometry, part segmentation, and joint parameters by rendering canonical Gaussians $\mathcal G^c$ into both states. The joint types are fixed to $\boldsymbol\kappa$, and the explicit parameters $\boldsymbol\Theta$ of Eq.~\eqref{eq:joint} are initialized at $\boldsymbol\Theta^{(0)}$. Let $T_k^{\pm1/2}$ denote the motion of joint $k$ by $\pm$ half of its motion magnitude, \ie, the angle $\pm\theta_k/2$ or the displacement $\pm s_k/2$, and $\mathbf R_k^{\pm1/2}$ its rotation. The base keeps the identity.

\paragraph{Mid-State Canonical Gaussians.}
Prior work typically builds the canonical Gaussians from the midpoints of corresponding Gaussians in the two states~\cite{liu2025building,yu2026part,wu2026reartgs++}. For a revolute part, such a midpoint lies on the chord of the rotation arc, at distance $r\cos(\theta_k/2)$ from the axis for a point at distance $r$, so the canonical part shrinks toward the axis. We therefore build the canonical model at the mid state of the initial joints. Each Gaussian of state 0 is moved by the half motion of its part $\ell_i=\ell_i^{(0)}$ under $\boldsymbol\Theta^{(0)}$,
\begin{equation}
    \boldsymbol\mu_i^c=T_{\ell_i}^{1/2}(\boldsymbol\mu_i^0),\qquad
    \boldsymbol\Sigma_i^c=\mathbf R_{\ell_i}^{1/2}\boldsymbol\Sigma_i^0\bigl(\mathbf R_{\ell_i}^{1/2}\bigr)^{\!\top},
    \label{eq:canonical}
\end{equation}
and keeps its opacity and appearance. Each movable part thus moves rigidly to the midpoint of its joint trajectory with its shape preserved. The canonical model is also consistent with the initial joints: $T^{-1/2}$ maps it back to $\mathcal G^0$, and $T^{1/2}$ maps it to the state 1 pose predicted by $\boldsymbol\Theta^{(0)}$, so the optimization starts close to both observed states.

\vspace{-5pt}
\paragraph{Articulated Rendering and Optimization.}
A center-based part field~\cite{liu2025building} assigns each canonical Gaussian to a part $k_i$ through a straight-through Gumbel softmax, and we fit it to $\boldsymbol\ell^{(0)}$ before optimization. A canonical Gaussian is rendered in state $s$ with
\begin{equation}
    \boldsymbol\mu_i^s=T_{k_i}^{s-1/2}(\boldsymbol\mu_i^c),\quad
    \boldsymbol\Sigma_i^s=\mathbf R_{k_i}^{s-1/2}\boldsymbol\Sigma_i^c\bigl(\mathbf R_{k_i}^{s-1/2}\bigr)^{\!\top},
    \label{eq:articulated_render}
\end{equation}
\ie, by $T^{-1/2}$ in state 0 and by $T^{1/2}$ in state 1. At each iteration, we sample a view from the combined training views of both states. For the sampled view, we jointly optimize $\mathcal G^c$, the part field, and $\boldsymbol\Theta$ with
\begin{equation}
    \mathcal L=\mathcal L_{\mathrm{rec}}+\lambda_{\mathrm{reg}}\mathcal L_{\mathrm{reg}}+\lambda_{\mathrm{pl}}\mathcal L_{\mathrm{pl}},
    \label{eq:refine_loss}
\end{equation}
where $\mathcal L_{\mathrm{rec}}$ is the RGB-D loss of Eq.~\eqref{eq:RGB-D_loss}, and $\mathcal L_{\mathrm{reg}}$ keeps each part center near the mean of its Gaussians~\cite{liu2025building}. $\mathcal L_{\mathrm{pl}}$ is a cross-entropy loss that supervises the part field with motion pseudo-labels. A canonical Gaussian receives the label of part $k$ when the motion of part $k$ alone moves it onto the reconstructed Gaussians of both states.

\section{Experiments}
\label{sec:experiments}

\subsection{Experimental Setup}
\label{sec:exp_setup}

\paragraph{Datasets \& Baselines.}
We compare GeArt with ArtGS~\cite{liu2025building}, ReArtGS++~\cite{wu2026reartgs++}, and GEAR~\cite{li2026gear} on three datasets: (1) PARIS~\cite{liu2023paris} contains ten synthetic and two real-world two-part objects. (2) DTA-Multi~\cite{weng2024neural} contains two synthetic multi-part objects. (3) ArtGS-Multi~\cite{liu2025building} contains five synthetic objects with three to six movable parts. 

\paragraph{Evaluation Metrics.}
We follow prior work to evaluate articulation accuracy and geometric reconstruction quality. For articulation, \textbf{Axis Ang} measures the sign-invariant angular error between predicted and ground-truth joint axes, while \textbf{Axis Pos} measures the shortest distance between revolute joint axes, normalized by object scale. \textbf{Part Motion} evaluates the motion magnitude error, measured in degrees for revolute joints and meters for prismatic joints. For geometry, we report the symmetric Chamfer distance (CD) between reconstructed and ground-truth surfaces, separately for static parts (CD-s), movable parts (CD-m), and the entire object (CD-w). 

\begin{table}[t!]
\centering
\begingroup
\caption{\textbf{Quantitative results on \textsc{\textbf{ArtGS-Multi}} and \textsc{\textbf{DTA-Multi}}}. Lower ($\downarrow$) is better across all metrics. \textbf{Best}, \underline{second}. ``-'' indicates objects containing only prismatic joints. N/A indicates that the results could not be evaluated due to incorrect joint-type predictions. Average values use the six common valid objects and exclude Table 25493.}
\vspace{-5pt}
\label{tab:multi_result}
\setlength{\tabcolsep}{2pt}
\resizebox{\linewidth}{!}{
\begin{tabular}{cc|ccccc|cc|c}
\hline
\multirow{2}{*}{Metric} & \multirow{2}{*}{Method} & \multicolumn{5}{c}{ArtGS-Multi} & \multicolumn{2}{|c}{DTA-Multi} & \multicolumn{1}{c}{} \\
& & Oven & \shortstack{Table\\25493} & \shortstack{Storage\\47648} & \shortstack{Table\\31249} & \shortstack{Storage\\45503} & Fridge & Storage & Average \\
\hline
\multirow{4}{*}{\shortstack{Axis\\Ang}} & ArtGS & \underline{0.04} & \underline{0.07} & 0.10 & 0.06 & 0.03 & \textbf{0.01} & \underline{0.05} & \underline{0.05} \\
& ReArtGS++ & 28.17 & 12.65 & 5.12 & 0.09 & 0.03 & 0.85 & \underline{0.05} & 5.72 \\
& GEAR & \textbf{0.02} & N/A & \underline{0.07} & \underline{0.04} & \underline{0.01} & 43.61 & \textbf{0.00} & 7.29 \\
& \our{\textbf{Ours}} & \our{\underline{0.04}} & \our{\textbf{0.03}} & \our{\textbf{0.05}} & \our{\textbf{0.03}} & \our{\textbf{0.00}} & \our{\underline{0.04}} & \our{\textbf{0.00}} & \our{\textbf{0.03}} \\
\hline
\multirow{4}{*}{\shortstack{Axis\\Pos}} & ArtGS & \underline{0.02} & - & \textbf{0.00} & \textbf{0.00} & \textbf{0.00} & \textbf{0.00} & \textbf{0.00} & \underline{0.00} \\
& ReArtGS++ & 1.56 & - & 0.90 & \textbf{0.00} & \textbf{0.00} & \underline{1.25} & \textbf{0.00} & 0.62 \\
& GEAR & \textbf{0.00} & - & \underline{0.01} & \textbf{0.00} & \underline{0.01} & 2.03 & \textbf{0.00} & 0.34 \\
& \our{\textbf{Ours}} & \our{\textbf{0.00}} & \our{-} & \our{\textbf{0.00}} & \our{\textbf{0.00}} & \our{\textbf{0.00}} & \our{\textbf{0.00}} & \our{\textbf{0.00}} & \our{\textbf{0.00}} \\
\hline
\multirow{4}{*}{\shortstack{Part\\Motion}} & ArtGS & 0.30 & \textbf{0.00} & 0.09 & \underline{0.04} & \textbf{0.02} & \textbf{0.03} & \textbf{0.00} & \underline{0.08} \\
& ReArtGS++ & 36.21 & \underline{0.07} & 15.27 & 0.05 & 0.04 & 13.60 & \underline{0.01} & 10.86 \\
& GEAR & \textbf{0.04} & N/A & \textbf{0.05} & \textbf{0.03} & \textbf{0.02} & 43.17 & \underline{0.01} & 7.22 \\
& \our{\textbf{Ours}} & \our{\underline{0.16}} & \our{\textbf{0.00}} & \our{\underline{0.07}} & \our{\textbf{0.03}} & \our{\underline{0.03}} & \our{\underline{0.04}} & \our{\underline{0.01}} & \our{\textbf{0.06}} \\
\hline
\multirow{4}{*}{CD-s} & ArtGS & \underline{1.10} & \textbf{0.72} & \textbf{0.68} & 1.05 & \underline{0.77} & \textbf{0.62} & 0.80 & \underline{0.84} \\
& ReArtGS++ & 9.19 & 1.15 & 1.91 & 2.07 & 2.09 & 5.02 & 4.67 & 4.16 \\
& GEAR & \textbf{0.73} & N/A & 1.84 & \underline{1.04} & \textbf{0.62} & \underline{0.67} & \textbf{0.70} & 0.93 \\
& \our{\textbf{Ours}} & \our{1.22} & \our{\underline{0.89}} & \our{\underline{0.77}} & \our{\textbf{0.86}} & \our{0.79} & \our{\textbf{0.62}} & \our{\underline{0.74}} & \our{\textbf{0.83}} \\
\hline
\multirow{4}{*}{CD-m} & ArtGS & \textbf{0.27} & \underline{6.47} & 3.61 & 1.39 & \underline{0.13} & \textbf{0.12} & \underline{0.23} & \underline{0.96} \\
& ReArtGS++ & 36.94 & 33.18 & 50.60 & 4.57 & 0.44 & 3.96 & 1.39 & 16.32 \\
& GEAR & 0.41 & N/A & \textbf{0.56} & \underline{0.87} & \textbf{0.11} & 119.11 & \underline{0.23} & 20.22 \\
& \our{\textbf{Ours}} & \our{\underline{0.31}} & \our{\textbf{0.24}} & \our{\underline{2.22}} & \our{\textbf{0.46}} & \our{0.17} & \our{\underline{0.63}} & \our{\textbf{0.22}} & \our{\textbf{0.67}} \\
\hline
\multirow{4}{*}{CD-w} & ArtGS & \underline{1.03} & \underline{0.75} & \underline{0.70} & 1.11 & 0.89 & \textbf{0.75} & \underline{0.96} & \underline{0.91} \\
& ReArtGS++ & 8.91 & 1.07 & 1.70 & 1.95 & 2.16 & 5.95 & 3.79 & 4.08 \\
& GEAR & \textbf{0.72} & N/A & 1.12 & \underline{1.02} & \textbf{0.77} & \underline{1.99} & 10.95 & 2.76 \\
& \our{\textbf{Ours}} & \our{1.14} & \our{\textbf{0.64}} & \our{\textbf{0.68}} & \our{\textbf{0.98}} & \our{\underline{0.88}} & \our{\textbf{0.75}} & \our{\textbf{0.88}} & \our{\textbf{0.89}} \\
\hline
\end{tabular}
}
\vspace{-5pt}
\endgroup
\end{table}

\subsection{Experimental Results}

\paragraph{Results on PARIS Dataset.}
Table~\ref{tab:paris_result} reports the quantitative comparison on the PARIS dataset. Since each object contains only one movable part, articulation estimation is relatively straightforward, and existing methods already achieve accurate results on most synthetic objects. GeArt achieves competitive performance in both joint estimation and geometric reconstruction. On real-world objects, GeArt achieves the lowest Part Motion errors and improves joint-axis estimation on Real-Storage. It also achieves the best static-part, movable-part, and whole-object reconstruction on Real-Storage. These results demonstrate that GeArt maintains competitive performance on simple two-part objects while achieving more stable and robust reconstruction of challenging real-world objects.

\begin{figure}[t]
  \centering
  \includegraphics[width=\linewidth]{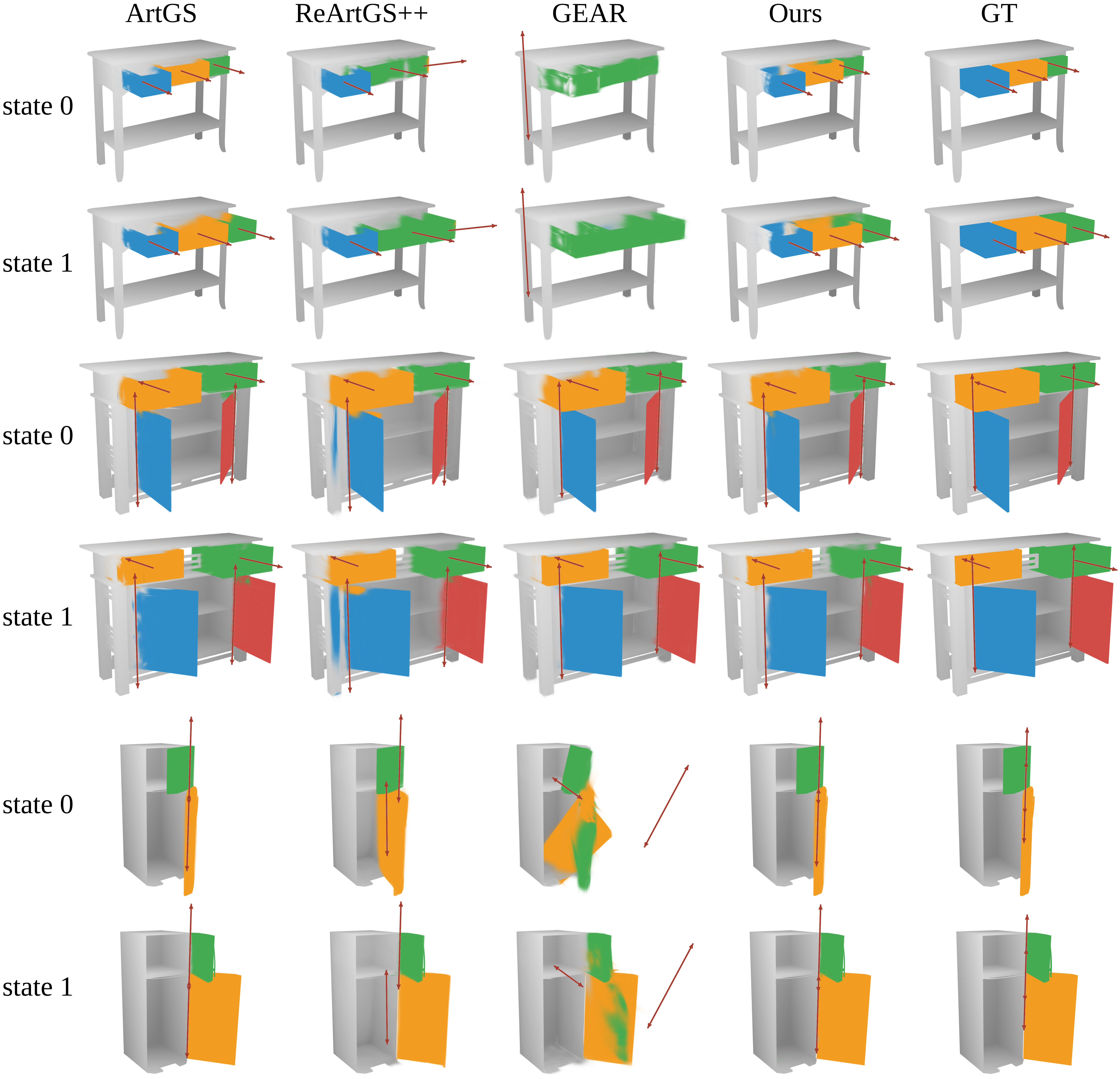}
  \caption{Qualitative results on complex multi-part articulated objects. Compared with other methods, our method achieves better agreement with the ground truth in reconstructed geometry, part segmentation, and joint parameters.}
  \label{fig:baseline_visual}
\end{figure}

\paragraph{Results on DTA-Multi and ArtGS-Multi Datasets.}
Table~\ref{tab:multi_result} presents the quantitative comparison on multi-part datasets. GeArt achieves the lowest average errors across all six metrics, demonstrating its effectiveness in both articulation estimation and geometric reconstruction. In particular, GeArt improves movable-part reconstruction on several challenging multi-part objects, where existing methods often produce larger geometric errors due to difficulties in obtaining reliable initialization before jointly optimizing geometry and articulation. These results highlight the advantage of global explanation-based articulation inference in providing reliable part assignments and joint estimates for subsequent refinement. Fig.~\ref{fig:baseline_visual} further shows that GeArt better separates adjacent parts and recovers accurate part geometry and joint axes across both states.

\subsection{Ablation Studies}
\label{sec:ablations}
We ablate the three main components of GeArt on DTA-Multi and ArtGS-Multi: surface-aware dynamic detection, the explanation score terms, and mid-state canonical initialization.

\begin{table}[t]
\centering
\begingroup
\setlength\extrarowheight{1pt}
\caption{\textbf{Ablation of dynamic Gaussian detection.} Comparison of detection precision, recall, F1 score, and off-surface rate. \textbf{Best}.}
\label{tab:ablation1}
\resizebox{\columnwidth}{!}{
\begin{tabular}{c c c c c c}
\toprule
\textbf{Recon.} & \textbf{Detection} & \textbf{Precision}$\uparrow$ & \textbf{Recall}$\uparrow$ & \textbf{F1}$\uparrow$ & \textbf{Off-surface (\%)}$\downarrow$ \\
\midrule
\multirow{2}{*}{RGB}
& Nearest-neighbor & 0.810 & \textbf{0.867} & 0.836 & 3.45 \\
    & Surface-aware    & 0.990 & 0.828 & \textbf{0.893} & \textbf{0.00} \\
\cmidrule(lr){1-6}
\multirow{2}{*}{RGB-D}
& Nearest-neighbor & 0.880 & 0.849 & 0.858 & 1.18 \\
& Surface-aware    & \textbf{0.996} & 0.819 & 0.889 & \textbf{0.00} \\
\bottomrule
\end{tabular}}
\endgroup
\end{table}

\paragraph{Effectiveness of Surface-Aware Dynamic Detection.}
We compare the cross-state nearest-neighbor detection used in prior works~\cite{liu2025building,wu2026reartgs++,yu2026part} with our surface-aware detection under two reconstruction settings: RGB-only reconstruction and RGB-D reconstruction with depth supervision and depth-based initialization. We report precision, recall, F1 score, and the proportion of detected Gaussians lying off the ground-truth surface.

\begin{figure}[t]
  \centering
  \includegraphics[width=\linewidth]{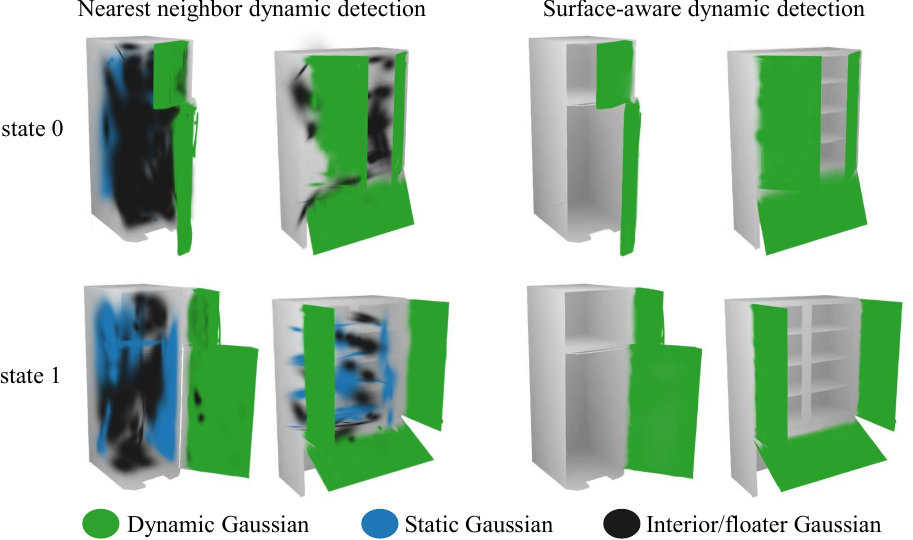}
  \caption{Qualitative comparison of dynamic Gaussian detection. Nearest-neighbor detection marks static, interior, and floating Gaussians as dynamic. Surface-aware detection keeps the Gaussians on the movable parts and removes these false detections in both states.}
  \label{fig:ablation1}
\end{figure}

As shown in Table~\ref{tab:ablation1}, surface-aware detection raises precision from 0.810 to 0.990 with RGB reconstruction and from 0.880 to 0.996 with RGB-D reconstruction, and it achieves the higher F1 score in both settings. It also removes every off-surface detection. Surface support and free-space evidence thus reject the static, interior, and floating Gaussians that nearest-neighbor detection marks as dynamic. The qualitative results in Fig.~\ref{fig:ablation1} further demonstrate that our method produces cleaner and more accurate dynamic regions aligned with the movable parts.

\paragraph{Effectiveness of the Explanation Score Terms.}
To assess the contribution of each term of the explanation score, we ablate the explanation score formulation in Eq.~\eqref{eq:explanation_score} using three variants: w/o geometry removes geometric agreement by setting $c(x,m)=\chi(x,m)$; w/o appearance removes appearance agreement ($\chi(\mathbf{x},m)=1$); w/o collision removes the collision penalty, giving $\pi(m)=1$.

\begin{table}[t]
\centering
\begingroup
\setlength\extrarowheight{1pt}
\caption{\textbf{Ablation of explanation score.} We evaluate the effectiveness of the geometry, appearance, and collision terms through quantitative evaluation of coarse part segmentation and articulation estimates before rendering-based refinement. \textbf{Best}.}
\label{tab:ablation2}
\setlength{\tabcolsep}{2.5pt}
\resizebox{\columnwidth}{!}{
\begin{tabular}{l c c c c c c}
\toprule
\textbf{Variant}
& \textbf{Axis Ang}$\downarrow$
& \textbf{Axis Pos}$\downarrow$
& \textbf{Part Motion}$\downarrow$
& \textbf{CD-s}$\downarrow$
& \textbf{CD-m}$\downarrow$
& \textbf{CD-w}$\downarrow$ \\
\midrule
Full
& \textbf{0.532}
& \textbf{0.031}
& \textbf{0.456}
& 2.360
& \textbf{14.401}
& \textbf{2.016} \\

w/o Geometry
& 9.259
& 3.173
& 6.720
& \textbf{2.342}
& 145.647
& 2.627 \\

w/o Appearance
& 7.187
& 0.083
& 1.177
& 2.641
& 52.691
& 2.148 \\

w/o Collision
& 0.631
& 0.035
& 0.692
& 2.658
& 18.088
& 2.392 \\

\bottomrule
\end{tabular}}
\endgroup
\end{table}

As shown in Table~\ref{tab:ablation2}, the full model achieves the lowest errors across all joint metrics, movable-part reconstruction, and whole-object reconstruction. Removing geometric agreement leads to the largest degradation in joint estimation and movable-part reconstruction accuracy. Excluding appearance consistency also increases these errors, highlighting its contribution to disambiguating geometrically similar motion hypotheses through appearance agreement. Likewise, removing the collision penalty increases both errors, confirming the importance of kinematic plausibility in hypothesis selection. Since these terms only affect global joint selection and dynamic Gaussian assignment, their removal primarily increases joint estimation and movable-part reconstruction errors, with relatively minor changes in static-part and whole-object reconstruction errors. These findings demonstrate that geometric and appearance agreement serve as the primary signals for joint hypothesis selection, while the kinematic plausibility further improves joint estimation accuracy.

\paragraph{Effectiveness of Mid-State Canonicalization.}
To evaluate the effectiveness of mid-state canonicalization, we compare two initialization strategies: \emph{Mid-point} and \emph{Mid-state}. Mid-point constructs canonical Gaussians by averaging corresponding Gaussian positions across the two states. Mid-state instead applies half of the estimated joint motion to the initial Gaussians, preserving the rigid geometry of each part. 

\begin{table}[t]
\centering
\begingroup
\caption{\textbf{Ablation of canonical representation.} Comparison of mid-point and mid-state representations for revolute and prismatic joints. \textbf{Best}.}
\label{tab:ablation3}
\setlength{\tabcolsep}{2.5pt}
\renewcommand{\arraystretch}{1.05}
\resizebox{\linewidth}{!}{%
\begin{tabular}{l l c c c c c c}
\toprule
\textbf{Joint Type} & \textbf{Method}
& \textbf{Axis Ang}$\downarrow$
& \textbf{Axis Pos}$\downarrow$
& \textbf{Part Motion}$\downarrow$
& \textbf{CD-s}$\downarrow$
& \textbf{CD-m}$\downarrow$
& \textbf{CD-w}$\downarrow$ \\
\midrule
\multirow{2}{*}{Revolute}
& Mid-point
& 0.039 & 0.008 & 0.150
& 0.839 & 0.157 & 0.886 \\
& Mid-state
& \textbf{0.021}
& \textbf{0.005}
& \textbf{0.050}
& \textbf{0.650}
& \textbf{0.105}
& \textbf{0.753} \\
\midrule
\multirow{2}{*}{Prismatic}
& Mid-point
& 0.048 & -
& \textbf{0.000}
& 0.862 & 4.750 & 0.872 \\
& Mid-state
& \textbf{0.047}
& -
& \textbf{0.000}
& \textbf{0.680}
& \textbf{2.493}
& \textbf{0.754} \\
\bottomrule
\end{tabular}%
}
\endgroup
\end{table}

As shown in Table~\ref{tab:ablation3}, mid-state improves both joint estimation and geometric reconstruction, with larger gains in joint estimation for revolute joints. For rotational motion, the arithmetic midpoint lies inside the rotation arc, shifting the canonical geometry toward the joint axis and distorting the part shape. For prismatic joints, both strategies produce equivalent midpoint positions under exact correspondences, resulting in nearly identical joint estimation errors. Mid-state also achieves lower reconstruction errors for prismatic joints. These results demonstrate the importance of motion-consistent canonical initialization, especially for revolute joints.
\section{Conclusion}
We presented GeArt, a global explanation-based framework for articulated object reconstruction from two-state multi-view RGB-D observations. GeArt first identifies moving surface Gaussians through surface-aware dynamic detection, providing reliable motion evidence for subsequent inference. It then generates diverse joint hypotheses and selects a consistent set that collectively explains both observed states based on geometry, appearance, and kinematic plausibility. The inferred structure provides initial part assignments, joint types, and joint parameters for rendering-based refinement, where mid-state canonicalization preserves rigid part geometry. Experiments on PARIS, ArtGS-Multi, and DTA-Multi demonstrate competitive performance on two-part objects and improved articulation estimation and geometric reconstruction on challenging multi-part objects. Ablation studies further validate the contributions of surface-aware detection, global articulation inference, and mid-state canonicalization. These results highlight the effectiveness of global articulation reasoning in providing reliable initialization for joint geometry and motion optimization.

{
    \small
    \bibliographystyle{ieeenat_fullname}
    \bibliography{main}

@String(CVPR= {IEEE Conf. Comput. Vis. Pattern Recog.})

@String(ICCV= {Int. Conf. Comput. Vis.})

@String(TOG= {ACM Trans. Graph.})

@String(ICLR = {Int. Conf. Learn. Represent.})

@String(CVPR  = {CVPR})

@String(ICCV  = {ICCV})

@String(TOG   = {ACM TOG})

@String(ICLR  = {ICLR})

@article{kerbl20233d,
  title={3d gaussian splatting for real-time radiance field rendering.},
  author={Kerbl, Bernhard and Kopanas, Georgios and Leimk{\"u}hler, Thomas and Drettakis, George and others},
  journal={ACM Trans. Graph.},
  volume={42},
  number={4},
  pages={139--1},
  year={2023}
}

@inproceedings{liu2025building,
  title={Building Interactable Replicas of Complex Articulated Objects via Gaussian Splatting.},
  author={Liu, Yu and Jia, Baoxiong and Lu, Ruijie and Ni, Junfeng and Zhu, Song-Chun and Huang, Siyuan},
  booktitle={ICLR},
  year={2025}
}

@inproceedings{shen2026gaussianart,
  title={Gaussianart: Unified modeling of geometry and motion for articulated objects},
  author={Shen, Licheng and Zhang, Saining and Li, Honghan and Yang, Peilin and Huang, Zihao and Zhang, Zongzheng and Zhao, Hao},
  booktitle={2026 International Conference on 3D Vision (3DV)},
  pages={384--395},
  year={2026},
  organization={IEEE}
}

@inproceedings{yu2026part,
  title={Part$^{2}$ GS: Part-aware Modeling of Articulated Objects using 3D Gaussian Splatting},
  author={Yu, Tianjiao and Shah, Vedant and Wahed, Muntasir and Shen, Ying and Nguyen, Kiet A and Lourentzou, Ismini},
  booktitle={Proceedings of the IEEE/CVF Conference on Computer Vision and Pattern Recognition},
  pages={18913--18923},
  year={2026}
}

@article{kim2025screwsplat,
  title={Screwsplat: An end-to-end method for articulated object recognition},
  author={Kim, Seungyeon and Ha, Junsu and Kim, Young Hun and Lee, Yonghyeon and Park, Frank C},
  journal={arXiv preprint arXiv:2508.02146},
  year={2025}
}

@inproceedings{wu2026reartgs++,
  title={Reartgs++: Generalizable articulation reconstruction with temporal geometry constraint via planar gaussian splatting},
  author={Wu, Di and Liu, Liu and Huang, Anran and Liu, Yuyan and Yu, Qiaojun and Liu, Shaofan and Song, Liangtu and Lu, Cewu},
  booktitle={Proceedings of the IEEE/CVF Conference on Computer Vision and Pattern Recognition},
  pages={1177--1186},
  year={2026}
}

@inproceedings{liu2023paris,
  title={Paris: Part-level reconstruction and motion analysis for articulated objects},
  author={Liu, Jiayi and Mahdavi-Amiri, Ali and Savva, Manolis},
  booktitle={Proceedings of the IEEE/CVF International Conference on Computer Vision},
  pages={352--363},
  year={2023}
}

@inproceedings{weng2024neural,
  title={Neural implicit representation for building digital twins of unknown articulated objects},
  author={Weng, Yijia and Wen, Bowen and Tremblay, Jonathan and Blukis, Valts and Fox, Dieter and Guibas, Leonidas and Birchfield, Stan},
  booktitle={Proceedings of the IEEE/CVF Conference on Computer Vision and Pattern Recognition},
  pages={3141--3150},
  year={2024}
}

@inproceedings{mu2021sdf,
  title={A-sdf: Learning disentangled signed distance functions for articulated shape representation},
  author={Mu, Jiteng and Qiu, Weichao and Kortylewski, Adam and Yuille, Alan and Vasconcelos, Nuno and Wang, Xiaolong},
  booktitle={2021 IEEE/CVF International Conference on Computer Vision (ICCV)},
  pages={12981--12991},
  year={2021},
  organization={IEEE}
}

@inproceedings{jiang2022ditto,
  title={Ditto: Building digital twins of articulated objects from interaction},
  author={Jiang, Zhenyu and Hsu, Cheng-Chun and Zhu, Yuke},
  booktitle={2022 IEEE/CVF Conference on Computer Vision and Pattern Recognition (CVPR)},
  pages={5606--5616},
  year={2022},
  organization={IEEE}
}

@inproceedings{tseng2022cla,
  title={Cla-nerf: Category-level articulated neural radiance field},
  author={Tseng, Wei-Cheng and Liao, Hung-Ju and Yen-Chen, Lin and Sun, Min},
  booktitle={2022 International Conference on Robotics and Automation (ICRA)},
  pages={8454--8460},
  year={2022},
  organization={IEEE}
}

@inproceedings{guo2025articulatedgs,
  title={Articulatedgs: Self-supervised digital twin modeling of articulated objects using 3d gaussian splatting},
  author={Guo, Junfu and Xin, Yu and Liu, Gaoyi and Xu, Kai and Liu, Ligang and Hu, Ruizhen},
  booktitle={2025 IEEE/CVF Conference on Computer Vision and Pattern Recognition (CVPR)},
  pages={27144--27153},
  year={2025},
  organization={IEEE}
}

@inproceedings{wu20244d,
  title={4d gaussian splatting for real-time dynamic scene rendering},
  author={Wu, Guanjun and Yi, Taoran and Fang, Jiemin and Xie, Lingxi and Zhang, Xiaopeng and Wei, Wei and Liu, Wenyu and Tian, Qi and Wang, Xinggang},
  booktitle={2024 IEEE/CVF Conference on Computer Vision and Pattern Recognition (CVPR)},
  pages={20310--20320},
  year={2024},
  organization={IEEE}
}

@inproceedings{geng2023gapartnet,
  title={Gapartnet: Cross-category domain-generalizable object perception and manipulation via generalizable and actionable parts},
  author={Geng, Haoran and Xu, Helin and Zhao, Chengyang and Xu, Chao and Yi, Li and Huang, Siyuan and Wang, He},
  booktitle={2023 IEEE/CVF Conference on Computer Vision and Pattern Recognition (CVPR)},
  pages={7081--7091},
  year={2023},
  organization={IEEE}
}

@article{wu2021vat,
  title={Vat-mart: Learning visual action trajectory proposals for manipulating 3d articulated objects},
  author={Wu, Ruihai and Zhao, Yan and Mo, Kaichun and Guo, Zizheng and Wang, Yian and Wu, Tianhao and Fan, Qingnan and Chen, Xuelin and Guibas, Leonidas and Dong, Hao},
  journal={arXiv preprint arXiv:2106.14440},
  year={2021}
}

@inproceedings{ma2023sim2real,
  title={Sim2real 2: Actively building explicit physics model for precise articulated object manipulation},
  author={Ma, Liqian and Meng, Jiaojiao and Liu, Shuntao and Chen, Weihang and Xu, Jing and Chen, Rui},
  booktitle={2023 IEEE International Conference on Robotics and Automation (ICRA)},
  pages={11698--11704},
  year={2023},
  organization={IEEE}
}

@inproceedings{lin2025splart,
  title={Splart: Articulation estimation and part-level reconstruction with 3d gaussian splatting},
  author={Lin, Shengjie and Fang, Jiading and Irshad, Muhammad Zubair and Guizilini, Vitor Campagnolo and Ambrus, Rares Andrei and Shakhnarovich, Greg and Walter, Matthew R},
  booktitle={2025 IEEE/CVF International Conference on Computer Vision (ICCV)},
  pages={8841--8851},
  year={2025},
  organization={IEEE}
}

@inproceedings{mo2021where2act,
  title={Where2act: From pixels to actions for articulated 3d objects},
  author={Mo, Kaichun and Guibas, Leonidas J and Mukadam, Mustafa and Gupta, Abhinav and Tulsiani, Shubham},
  booktitle={Proceedings of the IEEE/CVF International Conference on Computer Vision},
  pages={6813--6823},
  year={2021}
}

@article{xu2022universal,
  title={Universal manipulation policy network for articulated objects},
  author={Xu, Zhenjia and He, Zhanpeng and Song, Shuran},
  journal={IEEE robotics and automation letters},
  volume={7},
  number={2},
  pages={2447--2454},
  year={2022},
  publisher={IEEE}
}

@inproceedings{ester1996density,
  title={A density-based algorithm for discovering clusters in large spatial databases with noise},
  author={Ester, Martin and Kriegel, Hans-Peter and Sander, J{\"o}rg and Xu, Xiaowei},
  booktitle={Proceedings of the Second International Conference on Knowledge Discovery and Data Mining (KDD)},
  pages={226--231},
  year={1996}
}

@article{isack2012energy,
  title={Energy-based geometric multi-model fitting},
  author={Isack, Hossam and Boykov, Yuri},
  journal={International Journal of Computer Vision},
  volume={97},
  number={2},
  pages={123--147},
  year={2012}
}

@article{sturm2011probabilistic,
  title={A probabilistic framework for learning kinematic models of articulated objects},
  author={Sturm, J{\"u}rgen and Stachniss, Cyrill and Burgard, Wolfram},
  journal={Journal of Artificial Intelligence Research},
  volume={41},
  pages={477--526},
  year={2011}
}

@inproceedings{katz2013interactive,
  title={Interactive segmentation, tracking, and kinematic modeling of unknown 3d articulated objects},
  author={Katz, Dov and Kazemi, Moslem and Bagnell, J Andrew and Stentz, Anthony},
  booktitle={2013 IEEE International Conference on Robotics and Automation},
  pages={5003--5010},
  year={2013},
  organization={IEEE}
}

@article{hu2017learning,
  title={Learning to predict part mobility from a single static snapshot},
  author={Hu, Ruizhen and Li, Wenchao and Van Kaick, Oliver and Shamir, Ariel and Zhang, Hao and Huang, Hui},
  journal={ACM Transactions On Graphics (TOG)},
  volume={36},
  number={6},
  pages={1--13},
  year={2017},
  publisher={ACM, New York, NY, USA}
}

@inproceedings{wang2019shape2motion,
  title={Shape2motion: Joint analysis of motion parts and attributes from 3d shapes},
  author={Wang, Xiaogang and Zhou, Bin and Shi, Yahao and Chen, Xiaowu and Zhao, Qinping and Xu, Kai},
  booktitle={2019 IEEE/CVF Conference on Computer Vision and Pattern Recognition (CVPR)},
  pages={8868--8876},
  year={2019},
  organization={IEEE}
}

@article{yan2020rpm,
  title={RPM-Net: recurrent prediction of motion and parts from point cloud},
  author={Yan, Zihao and Hu, Ruizhen and Yan, Xingguang and Chen, Luanmin and Van Kaick, Oliver and Zhang, Hao and Huang, Hui},
  journal={arXiv preprint arXiv:2006.14865},
  year={2020}
}

@inproceedings{li2020category,
  title={Category-level articulated object pose estimation},
  author={Li, Xiaolong and Wang, He and Yi, Li and Guibas, Leonidas J and Abbott, A Lynn and Song, Shuran},
  booktitle={Proceedings of the IEEE/CVF conference on computer vision and pattern recognition},
  pages={3706--3715},
  year={2020}
}

@inproceedings{jain2021screwnet,
  title={Screwnet: Category-independent articulation model estimation from depth images using screw theory},
  author={Jain, Ajinkya and Lioutikov, Rudolf and Chuck, Caleb and Niekum, Scott},
  booktitle={2021 IEEE International Conference on Robotics and Automation (ICRA)},
  pages={13670--13677},
  year={2021},
  organization={IEEE}
}

@inproceedings{jiang2022opd,
  title={Opd: Single-view 3d openable part detection},
  author={Jiang, Hanxiao and Mao, Yongsen and Savva, Manolis and Chang, Angel X},
  booktitle={European Conference on Computer Vision},
  pages={410--426},
  year={2022},
  organization={Springer}
}

@article{yi2018deep,
  title={Deep part induction from articulated object pairs},
  author={Yi, Li and Huang, Haibin and Liu, Difan and Kalogerakis, Evangelos and Su, Hao and Guibas, Leonidas},
  journal={arXiv preprint arXiv:1809.07417},
  year={2018}
}

@inproceedings{kawana2022unsupervised,
  title={Unsupervised pose-aware part decomposition for man-made articulated objects},
  author={Kawana, Yuki and Mukuta, Yusuke and Harada, Tatsuya},
  booktitle={European Conference on Computer Vision},
  pages={558--575},
  year={2022},
  organization={Springer}
}

@inproceedings{heppert2023carto,
  title={Carto: Category and joint agnostic reconstruction of articulated objects},
  author={Heppert, Nick and Irshad, Muhammad Zubair and Zakharov, Sergey and Liu, Katherine and Ambrus, Rares Andrei and Bohg, Jeannette and Valada, Abhinav and Kollar, Thomas},
  booktitle={2023 IEEE/CVF Conference on Computer Vision and Pattern Recognition (CVPR)},
  pages={21201--21210},
  year={2023},
  organization={IEEE}
}

@inproceedings{luiten2024dynamic,
  title={Dynamic 3d gaussians: Tracking by persistent dynamic view synthesis},
  author={Luiten, Jonathon and Kopanas, Georgios and Leibe, Bastian and Ramanan, Deva},
  booktitle={2024 International Conference on 3D Vision (3DV)},
  pages={800--809},
  year={2024},
  organization={IEEE}
}

@inproceedings{yang2024deformable,
  title={Deformable 3d gaussians for high-fidelity monocular dynamic scene reconstruction},
  author={Yang, Ziyi and Gao, Xinyu and Zhou, Wen and Jiao, Shaohui and Zhang, Yuqing and Jin, Xiaogang},
  booktitle={2024 IEEE/CVF Conference on Computer Vision and Pattern Recognition (CVPR)},
  pages={20331--20341},
  year={2024},
  organization={IEEE}
}

@article{duisterhof2023md,
  title={Md-splatting: Learning metric deformation from 4d gaussians in highly deformable scenes},
  author={Duisterhof, Bardienus P and Mandi, Zhao and Yao, Yunchao and Liu, Jia-Wei and Shou, Mike Zheng and Song, Shuran and Ichnowski, Jeffrey},
  journal={arXiv preprint arXiv:2312.00583},
  volume={2},
  number={3},
  year={2023}
}

@inproceedings{huang2024sc,
  title={Sc-gs: Sparse-controlled gaussian splatting for editable dynamic scenes},
  author={Huang, Yi-Hua and Sun, Yang-Tian and Yang, Ziyi and Lyu, Xiaoyang and Cao, Yan-Pei and Qi, Xiaojuan},
  booktitle={2024 IEEE/CVF Conference on Computer Vision and Pattern Recognition (CVPR)},
  pages={4220--4230},
  year={2024},
  organization={IEEE}
}

@inproceedings{kratimenos2024dynmf,
  title={Dynmf: Neural motion factorization for real-time dynamic view synthesis with 3d gaussian splatting},
  author={Kratimenos, Agelos and Lei, Jiahui and Daniilidis, Kostas},
  booktitle={European Conference on Computer Vision},
  pages={252--269},
  year={2024},
  organization={Springer}
}

@article{wu2026reartgs,
  title={Reartgs: Reconstructing and generating articulated objects via 3d gaussian splatting with geometric and motion constraints},
  author={Wu, Di and Liu, Liu and Linli, Zhou and Huang, Anran and Song, Liangtu and Yu, Qiaojun and Wu, Qi and Lu, Cewu},
  journal={Advances in Neural Information Processing Systems},
  volume={38},
  pages={102889--102915},
  year={2026}
}

@article{li2026gear,
  title={GEAR: GEometry-motion Alternating Refinement for Articulated Object Modeling with Gaussian Splatting},
  author={Li, Jialin and Fu, Bin and Wang, Ruiping and Chen, Xilin},
  journal={arXiv preprint arXiv:2604.07728},
  year={2026}
}

@article{teitz1968heuristic,
  title={Heuristic methods for estimating the generalized vertex median of a weighted graph},
  author={Teitz, Michael B and Bart, Polly},
  journal={Operations research},
  volume={16},
  number={5},
  pages={955--961},
  year={1968},
  publisher={INFORMS}
}

@inproceedings{besl1992method,
  title={Method for registration of 3-D shapes},
  author={Besl, Paul J and McKay, Neil D},
  booktitle={Sensor fusion IV: control paradigms and data structures},
  volume={1611},
  pages={586--606},
  year={1992},
  organization={Spie}
}

@inproceedings{arya2001local,
  title={Local search heuristic for k-median and facility location problems},
  author={Arya, Vijay and Garg, Naveen and Khandekar, Rohit and Meyerson, Adam and Munagala, Kamesh and Pandit, Vinayaka},
  booktitle={Proceedings of the thirty-third annual ACM symposium on Theory of computing},
  pages={21--29},
  year={2001}
}
}

\clearpage

\clearpage
\setcounter{page}{1}
\maketitlesupplementary

\setcounter{section}{0}
\setcounter{table}{0}
\setcounter{figure}{0}
\setcounter{equation}{0}
\setcounter{algorithm}{0}
\renewcommand{\thesection}{\Alph{section}}
\renewcommand{\thetable}{S\arabic{table}}
\renewcommand{\thefigure}{S\arabic{figure}}
\renewcommand{\theequation}{S\arabic{equation}}
\renewcommand{\thealgorithm}{S\arabic{algorithm}}
\renewcommand{\theHsection}{S\Alph{section}}
\renewcommand{\theHtable}{S\arabic{table}}
\renewcommand{\theHfigure}{S\arabic{figure}}
\renewcommand{\theHequation}{S\arabic{equation}}
\def\theHalgorithm{S\arabic{algorithm}}

This supplementary material provides the implementation details and the additional experimental results that support the main paper. Section~\ref{sec:supp_impl} gives the implementation details of each component of GeArt, including the joint-set selection algorithm and the hyperparameters. Section~\ref{sec:supp_results} first evaluates the rendering quality on the ArtGS realscans dataset (Sec.~\ref{sec:supp_realscan}) and gives more qualitative comparisons (Sec.~\ref{sec:supp_qualitative}). It then ablates the joint-set solver, the dynamic detection, and the hypothesis model (Sec.~\ref{sec:supp_ablation}), and applies GeArt reconstructions to physical simulation (Sec.~\ref{sec:supp_simulation}).

\section{Implementation Details}
\label{sec:supp_impl}

This section gives the implementation details of each component of GeArt, including the algorithms and the parameters.

\subsection{Surface-Aware Dynamic Detection}
\paragraph{Per-state reconstruction.} We reconstruct each state with the coarse stage of ArtGS~\cite{liu2025building} and add depth initialization and depth supervision. For each state, we back-project the valid pixels of all training depth maps into 3D. We average the points in voxels of $1\%$ of the object diagonal and keep at most $5\times10^4$ points as initial Gaussian centers. Each state is optimized for $10^4$ iterations at half image resolution with Eq.~(3) of the main paper, where $\lambda_1=0.8$, $\lambda_{\mathrm{SSIM}}=0.2$, and $\lambda_D=0.5$. Table~\ref{tab:supp_hparams} lists detailed hyperparameters.

\paragraph{Dynamic detection.} The detection compares Gaussian depths with the input depth maps. Let $L$ be the diagonal of the bounding box of the Gaussian centers of both states. We set $\delta=0.015L$ and $\epsilon=0.02L$ in Eq.~(5) of the main paper. A view is occluded when the Gaussian lies more than $\delta$ behind the observed surface or projects outside the image. The condition $n_{\mathrm{free}}>\max(n_{\mathrm{surf}},1)$ requires at least two free-space views, and free-space views must outnumber surface views.

\begin{table}[t]
\centering
\caption{\textbf{Hyperparameters of GeArt.}}
\label{tab:supp_hparams}
\small
\setlength{\tabcolsep}{2pt}
\renewcommand{\arraystretch}{1.05}
\begin{tabular*}{\columnwidth}{@{\extracolsep{\fill}}lcc@{}}
\toprule
\textbf{Category} & \textbf{Parameter} & \textbf{Value} \\
\midrule
\multirow{2}{*}{Reconstruction}
 & $\lambda_1$ / $\lambda_{\mathrm{SSIM}}$ / $\lambda_D$ & 0.8 / 0.2 / 0.5 \\
 & Opacity binarization weight & 0.1 \\
\midrule
Detection & $\delta$ / $\epsilon$ & $0.015L$ / $0.02L$ \\
\midrule
\multirow{2}{*}{Joint inference}
 & $\tau$ & $0.01L$ \\
 & DBSCAN radius / MinPts & $\tau$ / 3 \\
\midrule
\multirow{5}{*}{Refinement}
 & Iterations / resolution & 20k / full \\
 & $\lambda_{\mathrm{reg}}$ / $\lambda_{\mathrm{pl}}$ & 0.1 / 0.1 \\
 & Temperature (cosine) & $1{\rightarrow}0.1$, 10k iters \\
 & Joint LR (angle / other) & $10^{-2}$ / $10^{-3}$ \\
 & Densification iters / interval & 500--15k / 100 \\
\bottomrule
\end{tabular*}
\end{table}

\subsection{Joint Hypothesis Generation}
\label{sec:supp_hypothesis}
We generate joint hypotheses from local fragments of dynamic Gaussians. Spatial clustering provides initial fragments, which may contain multiple moving parts or split one part into several fragments. We address these cases through sequential motion extraction and motion-based grouping.

\paragraph{Fragment generation.} DBSCAN~\cite{ester1996density} splits the dynamic Gaussians of each state into fragments, with radius $\tau=0.01L$ and three minimum samples. We keep the fragments with at least 30 Gaussians. The small radius keeps most fragments on one part, even when two movable parts touch or look alike. However, one fragment can still span two touching parts or break into several fragments, which have the same motion.  The later steps correct these fragments by motion.

\paragraph{Typed proposals.} A revolute joint has an axis direction, a pivot, and an angle. A prismatic joint has an axis direction and a displacement. Since the direction varies continuously over the sphere, we therefore take a few candidate directions from the object structure and fit the remaining parameters along each direction. For each fragment, the axis directions come from three sources. (1) The principal axes of the fragment give revolute directions. (2) Door hinges and drawer slides usually align with the edges of the static base. We therefore estimate the three principal directions of the base from its minimum-volume box and from the normals of its planar surfaces, and use them as candidate axes of both types. (3) A moving part appears as fragments in both states. We therefore register the fragment with trimmed ICP~\cite{besl1992method} to the fragments on the other state. A registered rotation above $3^\circ$ gives a revolute axis. Otherwise, the registered translation gives a prismatic direction. The displacement between the two centroids gives one more prismatic direction.

With a fixed direction, we reduce each joint to one free parameter and sweep it. A prismatic joint keeps only the displacement $s$, which we sweep over $[-0.6L, 0.6L]$ in steps of $0.25\tau$. A revolute joint keeps the angle $\theta$ and the pivot $\mathbf p$. The pivot follows from the angle and a partner, because the rotation must map the centroid $\mathbf c$ of the fragment onto the centroid $\mathbf c'$ of the partner. In the plane normal to the axis, $\mathbf p$ is the unique solution of $(\mathbf I-\mathbf R_\theta)\,\mathbf p=\mathbf c'-\mathbf R_\theta\mathbf c$ for $\theta\neq0$. For each partner, we therefore sweep only $\theta$ from $-150^\circ$ to $150^\circ$ in steps of $2^\circ$. Each sweep sample receives a score, the fraction of the moved Gaussians within $\tau$ of the other state, and each sweep keeps its two best samples. Trimmed ICP then refines these samples, restricted to the angle and pivot for a revolute joint and to the displacement for a prismatic joint.

\paragraph{Sequential motion extraction.} To handle a fragment that spans two touching parts, we use sequential extraction, which takes several hypotheses from one fragment in turn. We accept the hypothesis of a fragment when its score is at least 0.3 and when it moves at least 30 Gaussians within $\tau$ of the other state. We then remove these Gaussians and search again on the rest of the fragment, until a hypothesis fails either threshold or fewer than 30 Gaussians remain. A fragment that spans two parts therefore yields one hypothesis for each part. We run this extraction on the fragments of both states, because two parts can touch in one state and separate in the other. The fragments of state 1 also cover parts that are visible only in state 1. We express every hypothesis as a motion from state 0 to state 1.

\paragraph{Motion-based grouping.} To address a part that breaks into several fragments, we group the hypotheses by motion. The fragments of such a part yield hypotheses with nearly the same motion. Two hypotheses join one group when they move their combined support to positions that differ by less than $2\tau$ on average. Each connected component of this relation forms a group. In each group, we take the hypothesis with the largest $\pi(m)$ times the number of Gaussians in its support as the representative. Alg.~\ref{alg:supp_selection} selects the initial joints from these representatives, denoted by $\mathcal R\subseteq\mathcal H$.

\subsection{Global Joint-Set Selection}
\label{sec:supp_selection}

\begin{algorithm}[t]
\caption{Joint-set selection for Eq.~(11) of the main paper.}
\label{alg:supp_selection}
\begin{algorithmic}[1]
\Require hypothesis pool $\mathcal H$, group representatives $\mathcal R\subseteq\mathcal H$, number of parts $K$
\Ensure selected joints $\mathcal M^\star$
\State $\mathcal M\gets\emptyset$
\While{$|\mathcal M|<K-1$} \Comment{greedy construction}
  \State $\mathcal M\gets\mathcal M\cup\{\arg\min_{m\in\mathcal R\setminus\mathcal M}E(\mathcal M\cup\{m\})\}$
\EndWhile
\State $\mathcal M\gets\textsc{Interchange}(\mathcal M,\mathcal R)$ \Comment{initial set}
\For{$r=1,\dots,4$} \Comment{pool expansion}
  \State $(m'_1,\dots,m'_{K-1})\gets\mathcal M$
  \For{$k=1,\dots,K-1$}
    \State $\mathcal H\gets\mathcal H\cup\textsc{Fit}(\text{Gaussians assigned to }m'_k)$
  \EndFor
  \State $\mathcal M\gets\textsc{Interchange}(\mathcal M,\mathcal H)$
  \If{$\bar d(m_k,m'_k)<\tau$ for every $k$}
    \State \textbf{break}
  \EndIf
\EndFor
\State \Return $\mathcal M^\star\gets\mathcal M$
\end{algorithmic}
\end{algorithm}
Algorithm~\ref{alg:supp_selection} approximately minimizes Eq.~(11) using greedy construction followed by the interchange heuristic~\cite{teitz1968heuristic}. Greedy construction selects $K-1$ joints from $\mathcal R$, each time adding the candidate that minimizes the resulting $E$. Interchange then improves the set with two moves. A single replacement exchanges one selected joint for an unselected hypothesis. A pair replacement exchanges two selected joints for two unselected hypotheses at once~\cite{arya2001local}. It escapes local optima in which no single replacement lowers $E$ but a pair replacement does. Interchange applies improving single replacements until none remains, then tries pair replacements, and repeats until no move lowers $E$. Interchange can only select hypotheses in the pool, and the initial hypotheses are fit to single fragments. Pool expansion therefore fits new hypotheses to the Gaussians assigned to each selected joint, which can span several fragments, adds them to $\mathcal H$, and restarts interchange from the current set. We alternate expansion and interchange for at most four rounds. We stop early when $\bar d(m_k,m'_k)<\tau$ for every $k$. Here, $\bar d$ is the mean distance between the positions of the support of joint $k$ under the current and previous motions.

\subsection{Rendering-Based Refinement}
We optimize Eq.~(14) of the main paper for $2\times10^4$ iterations at full resolution. Before optimization, we fit the part field to $\boldsymbol\ell^{(0)}$ for 1{,}000 steps on the largest connected component of each movable part. The Gumbel temperature decreases from 1 to 0.1 with a cosine schedule over the first $10^4$ iterations, and the assignment is hard after this schedule. The regularization loss $\mathcal L_{\mathrm{reg}}$ has weight $\lambda_{\mathrm{reg}}=0.1$ and starts at iteration 3{,}000. $\mathcal L_{\mathrm{pl}}$ has weight $\lambda_{\mathrm{pl}}=0.1$ and starts at iteration 500. The pseudo-labels of all canonical Gaussians are recomputed every 100 iterations, and each iteration evaluates $\mathcal L_{\mathrm{pl}}$ on at most $2^{16}$ randomly sampled labeled Gaussians. The joint parameters use Adam with a learning rate of $10^{-2}$ for the angle and $10^{-3}$ for the other parameters. The rates stay constant for 5{,}000 iterations and then decay exponentially to $1\%$ of their initial values. Densification follows ArtGS from iteration 500 to 15{,}000 every 100 iterations with a gradient threshold of $10^{-3}$.

\section{Additional Experimental Results}
\label{sec:supp_results}

\subsection{Results on the ArtGS Real-Scan Set}
\label{sec:supp_realscan}
To evaluate GeArt on more real captures, we use the real-scan set of ArtGS~\cite{liu2025building}. It contains seven real-world articulated objects with RGB-D views of two states. The set provides no ground-truth geometry or joints, so we report the rendering quality on the training views.

\begin{table}[t]
\centering
\begingroup
\caption{\textbf{Additional results on the ArtGS realscan dataset.} Training-view reconstruction metrics are averaged over seven objects. We highlight \textbf{best} and \underline{second-best} results.}
\label{tab:supp_realscan}
\small
\setlength{\tabcolsep}{12pt}
\renewcommand{\arraystretch}{1.05}
\begin{tabular}{l c c c}
\toprule
\textbf{Method} & \textbf{PSNR}$\uparrow$ & \textbf{SSIM}$\uparrow$ & \textbf{LPIPS}$\downarrow$ \\
\midrule
ArtGS & \underline{26.150} & \underline{0.9455} & \underline{0.1022} \\
ReArtGS++ & 25.453 & 0.9373 & 0.1079 \\
GEAR & 20.122 & 0.8667 & 0.1449 \\
\midrule
\our{\textbf{GeArt (Ours)}} & \our{\textbf{27.579}} & \our{\textbf{0.9570}} & \our{\textbf{0.0957}} \\
\bottomrule
\end{tabular}
\endgroup
\end{table}

\begin{figure*}[t]
  \centering
  \includegraphics[width=0.85\textwidth]{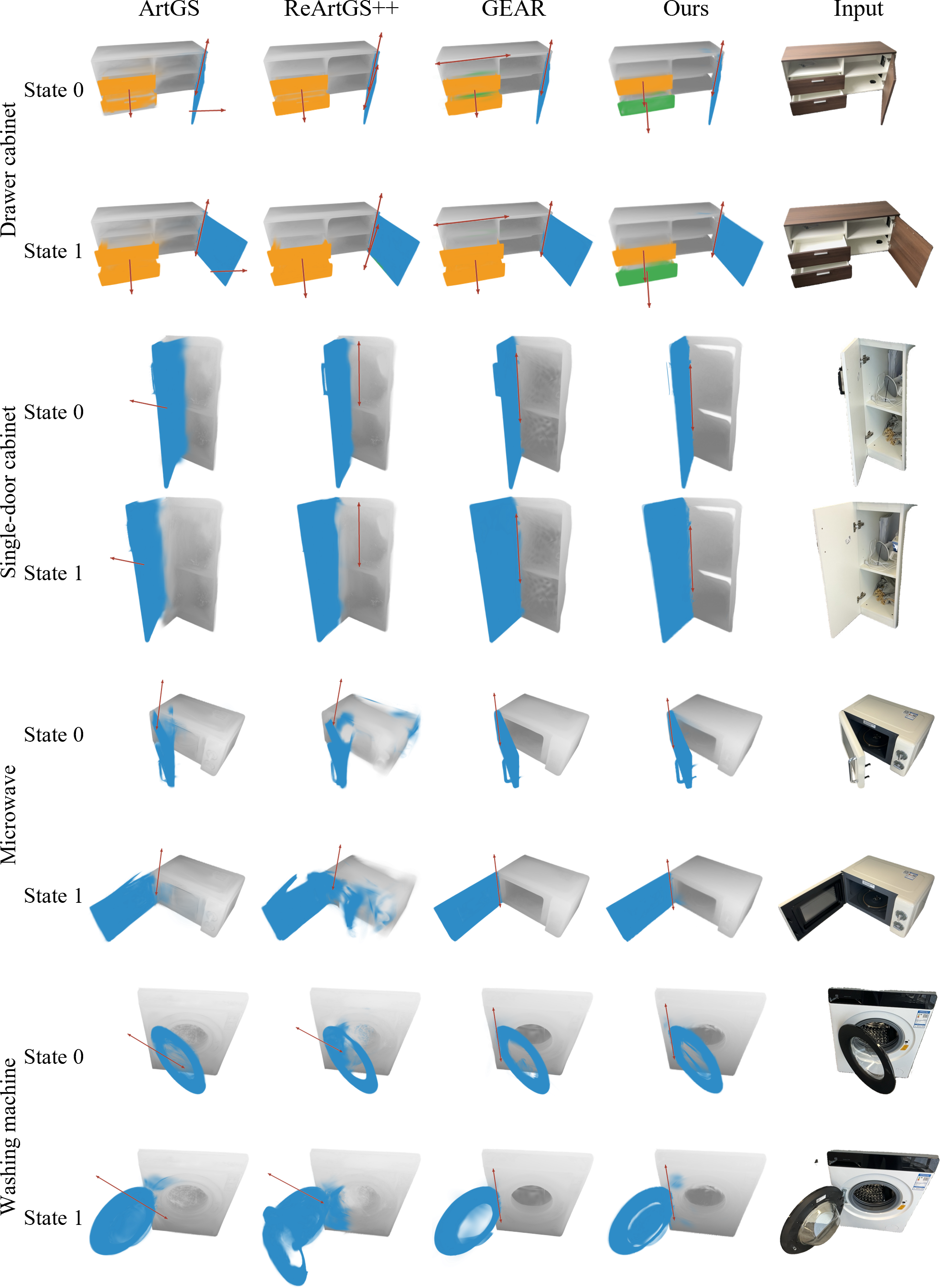}
  \caption{\textbf{Qualitative comparison on the ArtGS real-scan set.} On the drawer cabinet, ArtGS, ReArtGS++, and GEAR fail to separate two adjacent drawers with similar motion and geometry. In contrast, GeArt identifies them as distinct movable parts through a reliable canonical Gaussian initialization and reconstructs accurate geometry and motion. On the other objects, GeArt also recovers more accurate geometric boundaries and joint parameters than ArtGS and ReArtGS++.}
  \label{fig:supp_qual_realscan}
    \vspace{-10pt}
\end{figure*}

As shown in Table~\ref{tab:supp_realscan}, GeArt gives the best PSNR, SSIM, and LPIPS. It reaches a PSNR of 27.58\,dB, compared with 26.15\,dB for ArtGS. Figure~\ref{fig:supp_qual_realscan} shows four of these objects. On the drawer cabinet, GeArt separates the upper drawer, the lower drawer, and the door into three parts. The other methods assign both drawers mainly to one part. On the single-door cabinet and the microwave, GeArt gives a clearer boundary between the door and the body than ArtGS and ReArtGS++. These results demonstrate that GeArt separates adjacent movable parts more reliably and reconstructs real-world objects with higher rendering quality.

\subsection{Additional Qualitative Results}
\label{sec:supp_qualitative}
\begin{figure*}[t]
  \centering
  \includegraphics[width=0.85\textwidth]{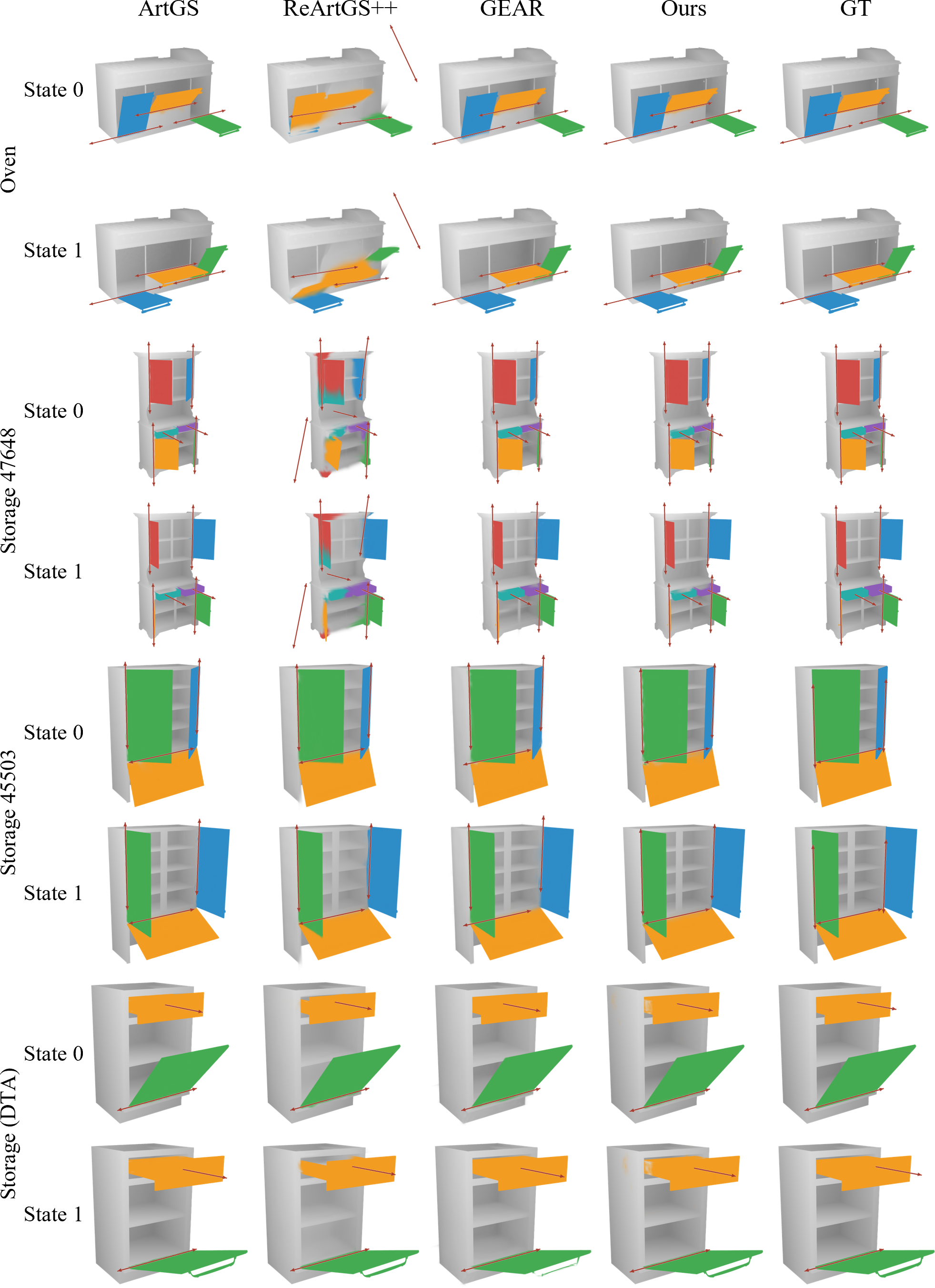}
  \caption{\textbf{Qualitative comparison on ArtGS-Multi and DTA-Multi.} On Oven and Storage 47648, ReArtGS++ mixes the movable parts and predicts axes far from the ground truth. In contrast, GeArt separates every movable part through a reliable canonical Gaussian initialization and recovers axes close to the ground truth. On the other objects, GeArt also recovers part segmentation and joint axes consistent with the ground truth.}
  \label{fig:supp_qual_multi}
  \vspace{-10pt}
\end{figure*}

\begin{figure*}[t]
  \centering
  \includegraphics[width=0.85\textwidth]{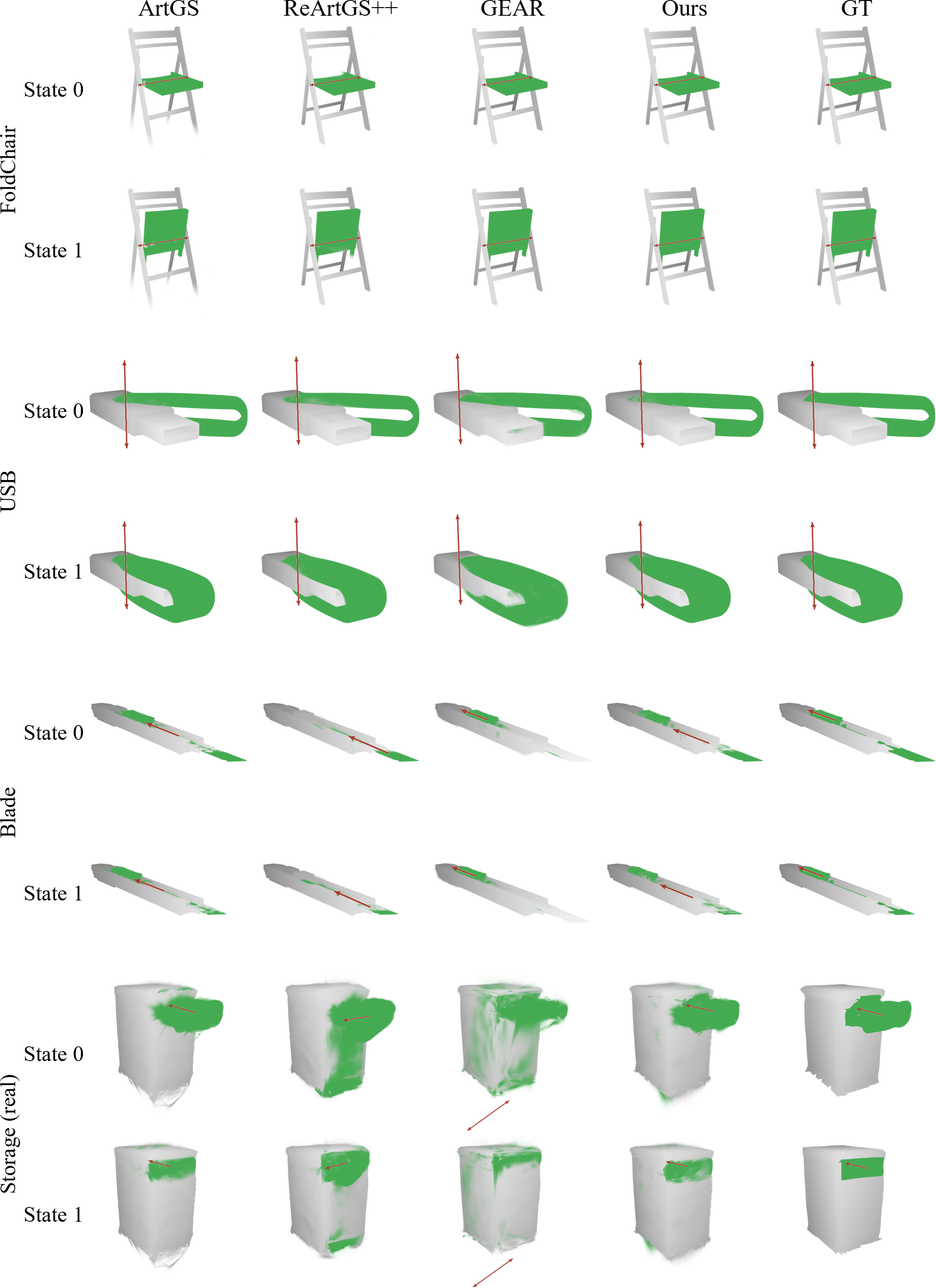}
  \caption{\textbf{Qualitative comparison on PARIS.} On Blade, ReArtGS++ and GEAR miss part of the sliding blade. On the real Storage, they spread the drawer label over the static base. In contrast, GeArt recovers the movable parts with far fewer mislabeled Gaussians than ReArtGS++ and GEAR, owing to its reliable canonical Gaussian initialization. On FoldChair and USB, GeArt also avoids the floating Gaussians of ArtGS and the blurred cover of GEAR.}
  \label{fig:supp_qual_paris}
  \vspace{-10pt}
\end{figure*}

Figures~\ref{fig:supp_qual_multi} and~\ref{fig:supp_qual_paris} show additional qualitative comparisons on ArtGS-Multi, DTA-Multi, and PARIS. On Oven and Storage 47648, ReArtGS++ mixes movable parts and predicts joint axes far from the ground truth. On Blade, ReArtGS++ and GEAR fail to reconstruct the complete sliding blade. On the real Storage, both methods incorrectly assign regions of the static base to the drawer. In contrast, GeArt separates the movable parts, preserves clear part boundaries, and recovers joint axes close to the ground truth. These results demonstrate that GeArt's reliable canonical Gaussian initialization supports accurate part segmentation and joint estimation on both synthetic and real objects.

\subsection{Additional Ablation Studies}
\label{sec:supp_ablation}
\paragraph{Effectiveness of solvers for global joint-Set selection}
\label{sec:supp_ablation_selection}
To evaluate the solver of Eq.~(11), we compare Alg.~\ref{alg:supp_selection} with three variants on the seven multi-part objects of DTA-Multi and ArtGS-Multi. All variants use the same hypothesis pool $\mathcal H$, the same representatives $\mathcal R$, and every other step of GeArt. They differ only in how they select the $K-1$ joints. \emph{Independent} ranks the representatives in $\mathcal R$ by their single-joint energy $E(\{m\})$ and keeps the $K-1$ lowest. \emph{Greedy} runs only the greedy construction over $\mathcal R$. \emph{Greedy + interchange} applies interchange over $\mathcal H$ to the greedy set, with single and pair replacements but without pool expansion. \emph{Full} runs the complete Alg.~\ref{alg:supp_selection} and is our method.

\begin{table*}[t]
\centering
\begingroup
\setlength\extrarowheight{1pt}
\caption{\textbf{Ablation of the joint-set selection.} Type error counts the predicted joints whose type differs from the matched ground-truth joint. Refinement keeps the joint types, so this count is the same before and after refinement. Joint errors average over the joints with the correct type. We highlight \textbf{best} results. Averages are over all seven objects}
\label{tab:supp_selection}
\setlength{\tabcolsep}{3pt}
\resizebox{\textwidth}{!}{%
\begin{tabular}{l c cccccc cccccc}
\toprule
& & \multicolumn{6}{c}{\textbf{Before refinement}} & \multicolumn{6}{c}{\textbf{After refinement}} \\
\cmidrule(lr){3-8}\cmidrule(lr){9-14}
\textbf{Selection variants} & \textbf{Type Error}$\downarrow$ & \textbf{Axis Ang}$\downarrow$ & \textbf{Axis Pos}$\downarrow$ & \textbf{Part Motion}$\downarrow$ & \textbf{CD-s}$\downarrow$ & \textbf{CD-m}$\downarrow$ & \textbf{CD-w}$\downarrow$ & \textbf{Axis Ang}$\downarrow$ & \textbf{Axis Pos}$\downarrow$ & \textbf{Part Motion}$\downarrow$ & \textbf{CD-s}$\downarrow$ & \textbf{CD-m}$\downarrow$ & \textbf{CD-w}$\downarrow$ \\
\midrule
Independent & 1 & 7.766 & 0.094 & 4.247 & 2.425 & 88.69 & 2.113 & 5.217 & 0.003 & 2.970 & 1.660 & 109.19 & 1.008 \\
Greedy & 1 & 6.876 & 0.083 & 1.489 & 2.351 & 52.64 & 2.202 & 6.223 & 0.003 & 0.126 & 0.971 & 55.53 & 0.929 \\
Greedy + interchange & 1 & 0.668 & 0.061 & 1.282 & \textbf{2.328} & 15.32 & 2.162 & 0.028 & 0.003 & 0.050 & 0.896 & 12.65 & 0.918 \\

Full (Ours) & \textbf{0} & \textbf{0.532} & \textbf{0.031} & \textbf{0.456} & 2.360 & \textbf{14.40} & \textbf{2.016} & \textbf{0.025} & \textbf{0.002} & \textbf{0.050} & \textbf{0.841} & \textbf{0.61} & \textbf{0.850} \\

\bottomrule
\end{tabular}%
}
\endgroup
\end{table*}

As shown in Table~\ref{tab:supp_selection}, the full solver is the only variant without a joint-type error. It gives the lowest joint errors before refinement and the lowest errors on all six metrics after refinement. Independent selection gives the largest CD-m before and after refinement, 88.69 and 109.19, because its joints do not compete for Gaussians. Greedy minimization of the set energy lowers this CD-m to 55.53. However, greedy construction never revises an earlier choice, and its Axis Ang after refinement remains 6.223$^\circ$, the largest among the variants. Interchange revises these choices and reduces this Axis Ang to 0.028$^\circ$. It still selects one joint of the wrong type. Refinement keeps the joint types, so this error remains and its CD-m stays at 12.65. Pool expansion corrects this joint type and lowers the CD-m to 0.61. Since the selection mainly affects the movable parts, CD-s and CD-w change much less than the joint errors and CD-m. These results demonstrate that our global joint-set selection, with the full minimization of Alg.~\ref{alg:supp_selection}, recovers the articulation structure, which refinement alone cannot correct.

\paragraph{Effectiveness of surface-aware dynamic detection}
\label{sec:supp_ablation_detection_final}
To comprehensively evaluate the effectiveness of surface-aware dynamic detection, we provide additional post-optimization results across the full set of articulation and geometry metrics. We run the complete pipeline with nearest-neighbor and surface-aware detection under the same two reconstruction settings.

\begin{table}[t]
\centering
\begingroup
\setlength\extrarowheight{1pt}
\caption{\textbf{Dynamic Gaussian detection and final accuracy.} Articulation and geometry errors after rendering-based refinement. We highlight \textbf{best} results.}
\label{tab:supp_detection_final}
\setlength{\tabcolsep}{2.5pt}
\resizebox{\columnwidth}{!}{%
\begin{tabular}{c c c c c c c c}
\toprule
\textbf{Recon.} & \textbf{Detection} & \textbf{Axis Ang}$\downarrow$ & \textbf{Axis Pos}$\downarrow$ & \textbf{Part Motion}$\downarrow$ & \textbf{CD-s}$\downarrow$ & \textbf{CD-m}$\downarrow$ & \textbf{CD-w}$\downarrow$ \\
\midrule
\multirow{2}{*}{RGB}
& Nearest-neighbor & 3.956 & 0.006 & 0.163 & 1.272 & 57.359 & 1.119 \\
& Surface-aware    & 3.958 & 0.007 & 0.187 & 1.233 & 29.343 & 1.098 \\
\cmidrule(lr){1-8}
\multirow{2}{*}{RGB-D}
& Nearest-neighbor & 6.068 & \textbf{0.002} & 0.094 & 0.896 & 50.802 & 0.855 \\
& Surface-aware (Ours) & \textbf{0.025} & \textbf{0.002} & \textbf{0.050} & \textbf{0.841} & \textbf{0.607} & \textbf{0.850} \\
\bottomrule
\end{tabular}%
}
\endgroup
\end{table}

As shown in Table~\ref{tab:supp_detection_final}, surface-aware detection with RGB-D reconstruction gives the lowest errors. With RGB-D reconstruction, it lowers Axis Ang from 6.068 to 0.025 and CD-m from 50.802 to 0.607. With RGB reconstruction, it halves CD-m from 57.359 to 29.343 and also lowers CD-s and CD-w. The static, interior, and floating Gaussians that nearest-neighbor detection marks as dynamic therefore degrade the final movable parts. These results demonstrate that the cleaner motion evidence of surface-aware detection carries over to the final joints and geometry.

\begin{figure*}[t]
  \centering
  \begin{tabular}{@{}c@{\hspace{4pt}}c@{}}
    \includegraphics[width=0.49\textwidth]{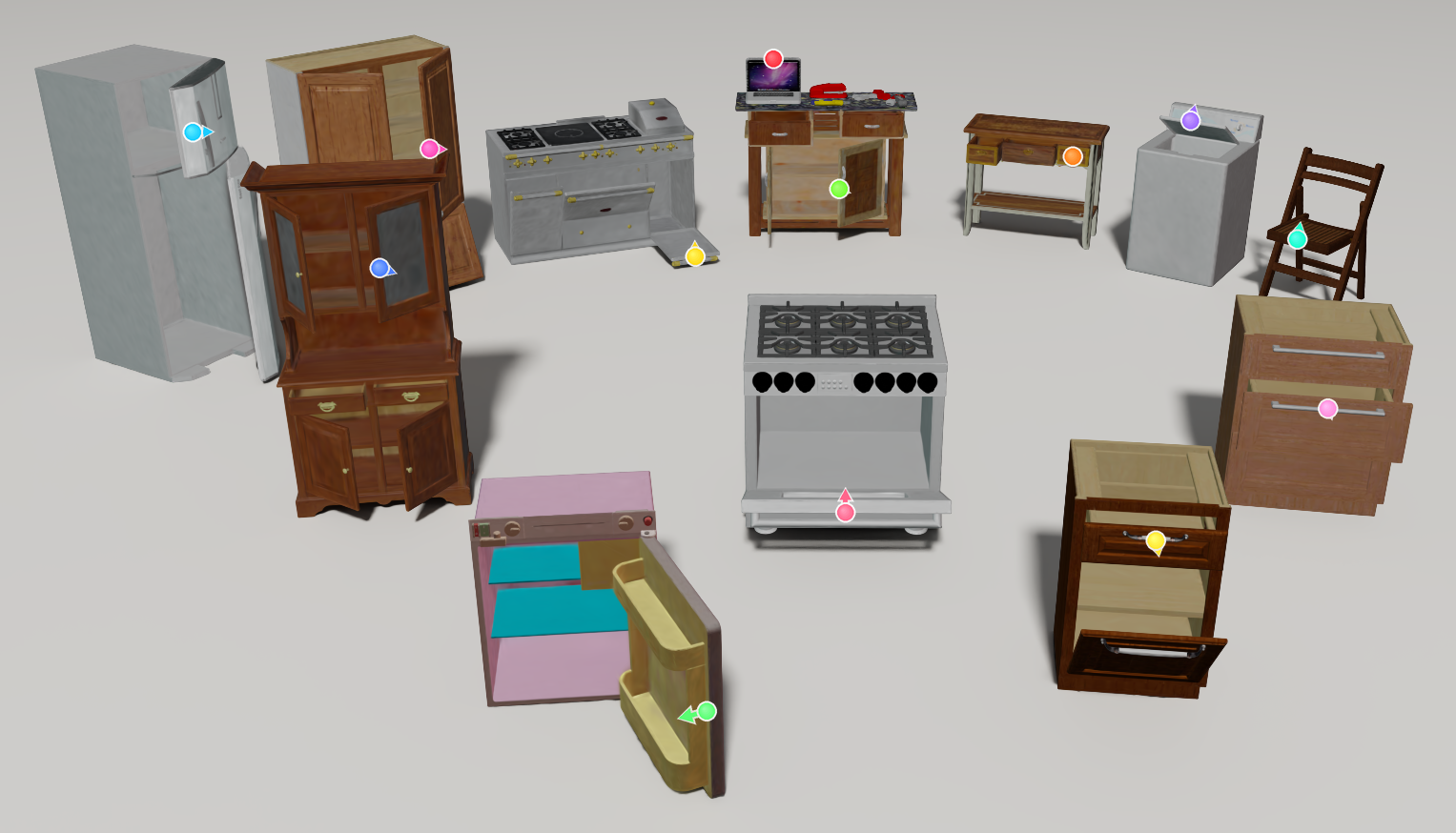} &
    \includegraphics[width=0.49\textwidth]{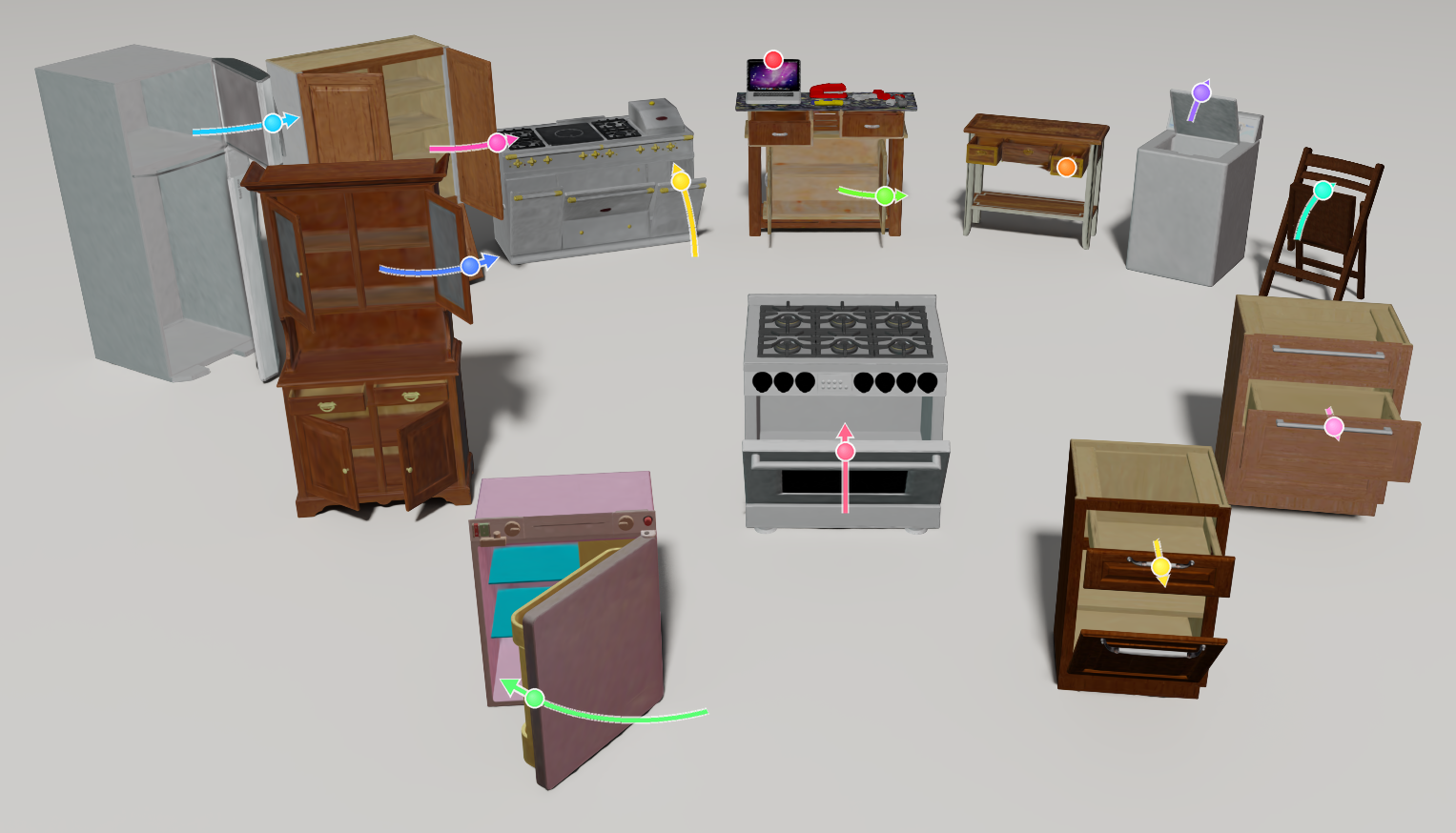} \\
    {\small (a) State 0} & {\small (b) State 1}
  \end{tabular}
  \vspace{-6pt}
  \caption{\textbf{Force-driven simulation of GeArt reconstructions in NVIDIA Isaac Sim.} (a) An external force, shown as a colored sphere and arrow, acts on one movable link of each object at state 0. (b) The configuration reached when the forces have driven the joints to state 1. The colored curves are the trajectories of the force points. Doors, lids, and the chair turn about the predicted hinges, and drawers move along the predicted sliding directions.}
  \label{fig:supp_isaac}
\end{figure*}

\paragraph{Structured joint hypotheses}
\label{sec:supp_ablation_hypothesis}
To evaluate the typed hypothesis model of Sec.~\ref{sec:supp_hypothesis}, we replace the motion fit of each fragment. Our \emph{1-DoF typed} model fixes the axis direction to the structural candidates and sweeps only the joint state. The \emph{3-DoF typed fit} fits a free rotation and a free translation to each fragment and keeps the better one. The \emph{6-DoF rigid fit} fits a free rigid motion with Kabsch registration and trimmed ICP, and then reads out the joint type and axis.

\begin{table}[t]
\centering
\begingroup
\setlength\extrarowheight{1pt}
\caption{\textbf{Ablation of the hypothesis model.} Results after refinement on the seven multi-part objects of DTA-Multi and ArtGS-Multi. Type Error counts the predicted joints whose type differs from the matched ground-truth joint. Refinement keeps the joint types, so this count is the same before refinement. Joint errors average over the joints with the correct type. We highlight \textbf{best} results.}
\label{tab:supp_hypothesis}
\setlength{\tabcolsep}{3pt}
\resizebox{\columnwidth}{!}{%
\begin{tabular}{l ccccccc}
\toprule
\textbf{Hypothesis model} & \textbf{Type Error}$\downarrow$ & \textbf{Axis Ang}$\downarrow$ & \textbf{Axis Pos}$\downarrow$ & \textbf{Part Motion}$\downarrow$ & \textbf{CD-s}$\downarrow$ & \textbf{CD-m}$\downarrow$ & \textbf{CD-w}$\downarrow$ \\
\midrule
3-DoF typed fit & 4 & 14.375 & 0.003 & 0.146 & 1.073 & 181.71 & 1.108 \\
6-DoF rigid fit & 3 & 6.029 & 0.066 & 2.254 & 4.611 & 72.66 & 1.251 \\
1-DoF typed (Ours) & \textbf{0} & \textbf{0.025} & \textbf{0.002} & \textbf{0.050} & \textbf{0.841} & \textbf{0.61} & \textbf{0.850} \\
\bottomrule
\end{tabular}%
}
\endgroup
\end{table}

As shown in Table~\ref{tab:supp_hypothesis}, our typed model is the only model without a joint-type error, and it gives the lowest errors. The 3-DoF and 6-DoF fits select 4 and 3 joints of the wrong type. The free fits also leave large axis errors after refinement, with an Axis Ang of 14.375$^\circ$ and 6.029$^\circ$ against 0.025$^\circ$ for ours. Their CD-m rises from 0.61 to 181.71 and 72.66. A free fit has more degrees of freedom than a physical joint, so on a small fragment it can absorb noise into a wrong motion. A typed hypothesis restricts this motion to a physical joint. These results demonstrate that typed hypotheses give more reliable joints than free motion fits.

\subsection{Application in Physical Simulation}
\label{sec:supp_simulation}
GeArt outputs explicit part geometry together with physical joint parameters, so its reconstructions can be used directly as interactive articulated assets. We demonstrate this in NVIDIA Isaac Sim with 17 GeArt reconstructions placed in a shared scene. For each object, the static part forms the root link and is fixed to the ground. Each movable part forms a child link attached to the root link by the estimated joint. A revolute joint takes the predicted hinge axis and pivot, and a prismatic joint takes the predicted sliding direction. The two observed states define the joint limits.

Instead of prescribing the joint states, we drive the objects with external forces. A force acts at a point on one movable link, such as the free edge of a door or the front of a drawer, and pulls this point toward a target that moves from state 0 to state 1. The physics engine simulates the link motion under the estimated joint constraints. As shown in Fig.~\ref{fig:supp_isaac}, the doors, the lids, and the folding chair turn about their predicted hinges, and the drawers move out along their predicted sliding directions. Every part comes to rest at its observed state 1. The force points trace circular arcs for revolute joints and straight segments for prismatic joints. These results demonstrate that GeArt reconstructions serve as simulation-ready articulated assets for downstream tasks such as robot manipulation and embodied interaction.

\end{document}